\documentclass[12pt]{article}
\usepackage{appendix}

\usepackage{graphicx}
\usepackage{amsfonts}
\usepackage{amsmath, amssymb,bbold}
\usepackage{algorithm}
\usepackage{algorithmic}
\usepackage{subfigure}
\usepackage{url}
\usepackage{epstopdf}
\usepackage{multirow}
\usepackage{cite}
\renewcommand{\baselinestretch}{1.1}

\newcommand{\va}{\boldsymbol{\alpha}}

\newcommand{\vbee}{\textbf{b}}

\newcommand{\ve}{\textbf{e}}

\newcommand{\vgee}{\textbf{g}}
\newcommand{\vh}{\textbf{h}}

\newcommand{\vr}{\textbf{r}}

\newcommand{\vu}{\textbf{u}}
\newcommand{\vv}{\textbf{v}}

\newcommand{\vx}{\textbf{x}}
\newcommand{\vy}{\textbf{y}}
\newcommand{\vz}{\textbf{z}}
\newcommand{\vzero}{\textbf{0}}

\newcommand{\mA}{\textbf{A}}

\newcommand{\mH}{\textbf{H}}
\newcommand{\mI}{\textbf{I}}

\newcommand{\mL}{\textbf{L}}

\newcommand{\mV}{\textbf{V}}
\newcommand{\mW}{\textbf{W}}

\def \RR {\mathbf{R}}

\newcommand{\argmax}[1]{{\hbox{\text{Arg}$\underset{#1}{\text{max}}\;$}}}
\newcommand{\argmin}[1]{{\hbox{\text{Arg}$\underset{#1}{\text{min}}\;$}}}

\newcommand{\gradient}{\nabla}

\newcommand{\casebegin}{\left\{\begin{array}{ll}}
	\newcommand{\caseend}{\end{array}\right.}

\newcommand{\ltriplebar}{\lvert\kern-0.9pt\lvert\kern-0.9pt\lvert}
\newcommand{\rtriplebar}{\rvert\kern-0.9pt\rvert\kern-0.9pt\rvert}

\newcommand{\bx}{\textbf{x}}

\newcommand{\bA}{\textbf{A}}
\newcommand{\bB}{\textbf{B}}

\newcommand{\bD}{\textbf{D}}

\newcommand{\bSigma}{\mathbf{\Sigma}}
\newcommand{\bH}{\textbf{H}}
\newcommand{\bI}{\textbf{I}}

\newcommand{\bK}{\textbf{K}}
\newcommand{\bL}{\textbf{L}}

\newcommand{\bR}{\textbf{R}}

\newcommand{\bW}{\textbf{W}}

\title{{ \it The Little Engine that Could} \\ {Regularization by Denoising (RED)}}
\date{}
\author{Yaniv Romano\footnotemark[1]\
	\and Michael Elad\footnotemark[3]\
	\and Peyman Milanfar\footnotemark[3]}

\begin{document}
\maketitle

\renewcommand{\thefootnote}{\fnsymbol{footnote}}
\footnotetext[1]{The Electrical Engineering Department, The Technion - Israel Institute of Technology.}
\footnotetext[3]{Google Research, Mountain View, California.}

\renewcommand{\thefootnote}{\arabic{footnote}}

\begin{abstract}
Removal of noise from an image is an extensively studied problem in image processing. Indeed, the recent advent of sophisticated and highly effective denoising algorithms lead some to believe that existing methods are touching the ceiling in terms of noise removal performance. Can we leverage this impressive achievement to treat other tasks in image processing? Recent work has answered this question positively, in the form of the Plug-and-Play Prior ($P^3$) method, showing that any inverse problem can be handled by sequentially applying image denoising steps. This relies heavily on the ADMM optimization technique in order to obtain this chained denoising interpretation. \\ \indent Is this the only way in which tasks in image processing can exploit the {\em image denoising engine}? In this paper we provide an alternative, more powerful and more flexible framework for achieving the same goal. As opposed to the $P^3$ method, we offer Regularization by Denoising (RED): using the denoising engine in defining the regularization of the inverse problem. We propose an explicit image-adaptive Laplacian-based regularization functional, making the overall objective functional clearer and better defined. With a complete flexibility to choose the iterative optimization procedure for minimizing the above functional, RED is capable of incorporating any image denoising algorithm, treat general inverse problems very effectively, and is guaranteed to converge to the globally optimal result. We test this approach and demonstrate state-of-the-art results in the image deblurring and super-resolution problems.
\end{abstract}

\noindent {\bf keywords:} Image Denoising Engine, Plug-and-Play Prior, Laplacian Regularization, Inverse Problems.

%\renewcommand{\baselinestretch}{1.1}
%\pagebreak

%%%%%%%%%%%%%%%%%%%%%%%%%%%%%%%%%%%%%%%%%%%%%%%%%%%%%%%%%%%%%%%%%%%%%%%%%%%%%%%%
%%%%%%%%%%%%%%%%%%%%%%%%%%%%%%%%%%%%%%%%%%%%%%%%%%%%%%%%%%%%%%%%%%%%%%%%%%%%%%%%

\section{Introduction}

We open this paper with a bold and possibly controversial statement: {\em To a large extent, removal of zero-mean white additive Gaussian noise from an image is a solved problem in image processing}.

\noindent Before justifying this statement, let us describe the basic building block that will be the star of this paper: {\em the image denoising engine}. From the humble linear Gaussian filter to the recently developed state-of-the-art methods using convolutional neural networks, there is no shortage of denoising approaches. In fact, these algorithms are so widely varied in their definition and underlying structure that a concise description will need to be made carefully. Our story begins with an image $\vx$ corrupted by zero-mean white additive Gaussian noise,
\begin{equation}\label{eqn:model}
\vy = \vx + \ve.
\end{equation}
In our notation, we consider an image as a vector of length $n$ (after lexicographic ordering). In the above description, the noise vector is normally distributed, $\ve \sim {\cal N}\left( \underline{0},\sigma^2 \bI\right)$. In the most general terms, the image denoising engine is a function $f:[0,255]^n \longrightarrow [0,255]^n$ that maps an image $\vy$ to another image of the same size ${\widehat \vx}=f(\vy)$, with the hope to get as close as possible to the original image $\vx$. Ideally, such functions operate on the input image $\vy$ to remove the deleterious effect of the noise while maintaining edges and textures beneath.

The claim made above about the denoising problem being solved is based on the availability of algorithms proposed in the past decade that can treat this task extremely effectively and stably, getting very impressive results, which also tend to be quite concentrated (see for example the work reported in \cite{NLM3,KSVDden,BM3D,KerReg,Mairal,GMM1,CSR,NCSR,ChattPeyman,NNden,Lebrun1,Lebrun2,DDID,Hossein,WNNM,NL-GSM,Morel}). Indeed, these documented results have led researchers to the educated guess that these methods are getting very close to the optimally possible denoising performance \cite{PeymanDenoisingDead,LevinNadler1,LevinNadler2}. This aligns well with the unspoken observation in our community in recent years that investing more work to improve image denoising algorithms seems to lead to diminishing returns.

While the above may suggest that work on denoising algorithms is turning to a dead-end avenue, a new opportunity emerges from this trend: Seeking ways to leverage the vast progress made on the image denoising front in order to treat other tasks in image processing, bringing their solutions to new heights. One natural path towards addressing this goal is to take an existing and well-performing denoising algorithm, and generalize it to handle a new problem. This has been the logical path that has led to contributions such as \cite{DenGen1,DenGen2,DenGen3,DenGen4,DenGen5,DenGen6}, and many others. These papers, and others like them, offer an exhaustive manual adaptation of existing denoising algorithms, carefully re-tailored to handle specific alternative problems. This line of work, while often successful, is quite limited, as it offers no flexibility and no general scheme  for diverting image denoising engines to treat new image processing tasks.

Could one offer a more systematic way to exploit the abundance of high-performing image-denoising algorithms to treat a much broader family of problems? The recent work by Venkatakrishnan, Bouman and Wohlberg provides a positive and tantalizing answer to this question, in the form of the Plug-and-Play Prior ($P^3$) method \cite{P3,Follows2,Follows1,Follows11}. This technique builds on the use of an implicit prior for regularizing general inverse problems. When handling the obtained optimization task via the ADMM optimization scheme \cite{ADMM1,ADMM2,ADMM3}, the overall problem decomposes into a sequence of image denoising tasks, coupled with simpler $L_2$-regularized inverse problems that are much easier to handle.

While the $P^3$ scheme may sound like the perfect answer to our prayers, reality is somewhat more complicated. First, this method is not always accompanied by a clear definition of the objective function, since the regularization being effectively used is only implicit, implied by the denoising algorithm. Indeed, it is not clear at all that there is an underlying objective function behind the $P^3$ scheme, if arbitrary denoising engines are used \cite{Follows1}. Second, parameter tuning of the ADMM scheme is a delicate matter, and especially so under a non-provable convergence regime, as is the case when using sophisticated denoising algorithms. Third, being intimately coupled with the ADMM, the $P^3$ scheme does not offer easy and flexible ways of replacing the iterative procedure. Because of these reasons, the $P^3$ scheme is not a turn-key tool, nor is it free from emotional-involvement. Nevertheless, the $P^3$ method has drawn much attention (e.g., \cite{Follows1,Follows11,Follows2,Follows3,Follows4,Follows5,Follows6,PPP-Convergence}), and rightfully so, as it offers a clear path towards harnessing a given image denoising engine for treating more general inverse problems, just as described above.

Is there a more general alternative to the $P^3$ method that could be simpler and more stable? This paper puts forward such a framework, offering a systematic use of such denoising engines for regularization of inverse problems. We term the proposed method ``Regularization by Denoising" (RED), relying on a general structured smoothness penalty term harnessed to regularize any desired inverse problem. More specifically, the regularization term we propose in this work is of the following
\begin{equation}
\rho(\vx) = \frac{1}{2} \: \vx^T\left[\vx-f(\vx)\right],
\end{equation}
in which the denoising engine itself is applied on the candidate image $\vx$, and the penalty induced is proportional to the inner-product between this image and its denoising residual. This defined smoothness regularization is effectively using an image-adaptive Laplacian, which in turn draws its definition from the arbitrary image denoising engine of choice, $f(\cdot)$. Surprisingly, under mild assumptions on $f(\cdot)$, it is shown that the gradient of the regularization term is manageable, given as the denoising residual, $\vx -f(\vx)$. Therefore, armed with this regularization expression, we show that any inverse problem can be handled while calling the denoising engine iteratively.

RED, the newly proposed framework, is much more flexible in the choice of the optimization method to use, not being tightly coupled to one specific technique, as in the case of the $P^3$ scheme (relying on ADMM). Another key difference w.r.t. the $P^3$ method is that our adaptive Laplacian-based regularization functional is explicit, making the overall Bayesian objective function clearer and better defined.  RED is capable of incorporating any image denoising algorithm, and can treat general inverse problems very effectively, while resulting in an overall algorithm with very simple structure.

An important advantage of RED over the $P^3$ scheme is the flexibility with which one can choose the denoising engine $f(\cdot)$ to plug in the regularization term. While most of the discussion in this paper keeps focusing on White Gaussian Noise (WGN) removal, RED can actually deploy almost any denoising engine. Indeed, we define a set of two mild conditions that $f(\cdot)$ should satisfy, and show that many known denoising methods obey these properties. As an example, in our experiments we show how the median filter can become an effective regularizer. Last but not least, we show that the defined regularization term is a convex function, implying that in most cases, in which the log-likelihood term is convex too, the proposed algorithms are guaranteed to converge to a global optimum solution. We demonstrate this scheme, showing state-of-the-art results in image deblurring and single image super-resolution.

This paper is organized as follows: In the next section we present the background material for this work, discussing the general form of inverse problems as optimization tasks, and presenting the Plug-and-Play Prior scheme. Section \ref{sec:Engine} focuses on the image denoising engine, defining it and its properties clearly, so as to enable its use in the proposed Laplacian paradigm. Section \ref{sec:Algorithm} serves the main part of this work -- introducing RED: a new way to use an image denoising engine to handle general structured inverse problems. In Section \ref{sec:Analysis} we analyze the proposed scheme, discussing convexity, an alternative formulation, and a qualitative comparison to the $P^3$ scheme. Results on the image deblurring and single-image super-resolution problems are brought in Section \ref{sec:Results}, demonstrating the strength of the proposed scheme. We conclude the paper in Section \ref{sec:Conclusions} with a summary of the open questions that we identify for future work.

%%%%%%%%%%%%%%%%%%%%%%%%%%%%%%%%%%%%%%%%%%%%%%%%%%%%%%%%%%%%%%%%%%%%%%%%%%%%%%%%
%%%%%%%%%%%%%%%%%%%%%%%%%%%%%%%%%%%%%%%%%%%%%%%%%%%%%%%%%%%%%%%%%%%%%%%%%%%%%%%%

\section{Preliminaries}\label{sec:Background}

In this section we provide background material that serves as the foundation to this work. We start by presenting the breed of optimization tasks we will work on throughout the paper for handling the inverse problems of interest. We then introduce the $P^3$ method and discuss its merits and weaknesses.

\subsection{Inverse Problems as Optimization Tasks}
Bayesian estimation of an unknown image $\vx$ given its measured version $\vy$ uses the posterior conditional probability, $P(\vx \vert \vy)$, in order to infer $\vx$. The most popular estimator in this regime is the Maximum aposteriori Probability (MAP), which chooses the mode ($\vx$ for which the maximum probability is obtained) of the posterior. Using Bayes' rule, this implies that the estimation task is turned into an optimization problem of the form
\begin{eqnarray}\label{eqn:BayesianMAP}
{\widehat \vx}_{MAP} = \argmax{\vx} P(\vx \vert \vy) & = & \argmax{\vx} \frac{P(\vy \vert \vx) P(\vx)}{P(\vy)}
= \argmax{\vx} P(\vy \vert \vx) P(\vx) \\ \nonumber
& = & \argmin{\vx} -\log\{P(\vy \vert \vx)\} -\log\{ P(\vx)\}.
\end{eqnarray}
In the above derivations we exploited the fact that $P(\vy)$ is not a function of $\vx$ and thus can be omitted. We also used the fact that the $-\log$ function is monotonic decreasing, turning the maximization into a minimization problem.

The term $-\log\{P(\vy \vert \vx)\}$ is known as the log-likelihood term, and it encapsulates the probabilistic relationship between the desired image $\vx$ and the measurements $\vy$, under the assumption that the desired image is known. We shall rewrite this term as
\begin{eqnarray}
\ell(\vy, \vx) = - \log\{P(\vy \vert \vx)\}.
\end{eqnarray}
As a classic example for the log-likelihood that will accompany us throughout this paper, the expression $\ell(\vy, \vx) = \frac{1}{2\sigma^2}\|\bH \vx - \vy\|_2^2$ refers to the case of $\vy = \bH \vx+ \ve $, where $\bH$ is a linear degradation operator and $\ve$ is white Gaussian noise contamination of variance $\sigma^2$. Naturally, if the noise distribution changes, we depart form the comfortable $L_2$ form.

The second term in Equation (\ref{eqn:BayesianMAP}), $-\log\{ P(\vx)\}$, refers to the {\em prior}, bringing in the influence of the statistical nature of the unknown. This term is also referred to as the regularization, as it helps in better conditioning the overall optimization task in cases where the likelihood alone cannot lead to a unique or stable solution. We shall rewrite this term as
\begin{eqnarray}
\lambda \rho(\vx) = - \log\{P( \vx)\},
\end{eqnarray}
where $\lambda$ is a scalar that encapsulates the confidence in this term.

What is $\rho(\vx)$ and how is it chosen? This is the holy grail of image processing, with a progressive advancement over the years of better modeling the image statistics and leveraging this for handling various tasks in image processing. Indeed, one could claim that almost everything done in our field surrounds this quest for choosing a proper prior, from the early smoothness prior $\rho(\vx) =\lambda \vx^T \bL \vx $ using the classic Laplacian \cite{LaplacianReg}, through total variation \cite{TV} and wavelet sparsity \cite{Wavelet}, all the way to recent proposals based on patch-based GMM \cite{GMM1,GMM2} and sparse-representation modeling \cite{SparseModel}. Interestingly, the work we report here builds on the surprising comeback of the Laplacian regularization in a much more sophisticated form, as reported in \cite{LapReg1,LapReg2,LapReg3,LapReg4,Milanfar_SPM,LapReg5,LapReg6,LapReg7,LapReg8,LapReg9}.

Armed with a clear definition of the relation between the measurements and the unknown, and with a trusted prior, the MAP estimation boils down to the optimization problem of the form
\begin{eqnarray}\label{eqn:BayesianMAP0}
{\widehat \vx}_{MAP} = \argmin{\vx} \ell(\vy , \vx) + \lambda \rho(\vx).
\end{eqnarray}
This defines a wide family of inverse problems that we aim to address in this work, which includes tasks such as denoising, deblurring, super-resolution, demosaicing, tomographic reconstruction, optical-flow estimation, segmentation, and many other problems. The randomness in these problems is typically due to noise contamination of the measurements, and this could  be Gaussian, Laplacian, Gamma-distributed, Poisson, and other noise models.

%%%%%%%%%%%%%%%%%%%%%%%%%%%%%%%%%%%%%%%%%%%%%%%%%%%%%%%%%%%%%%%%%%%%%%%%%%%%%%%%
%%%%%%%%%%%%%%%%%%%%%%%%%%%%%%%%%%%%%%%%%%%%%%%%%%%%%%%%%%%%%%%%%%%%%%%%%%%%%%%%

\subsection{The Plug-and-Play Prior ($P^3$) Approach}

For completeness of this exposition, we briefly review the $P^3$ approach. Aiming to solve the problem posed in Equation (\ref{eqn:BayesianMAP0}), the ADMM technique \cite{ADMM1,ADMM2,ADMM3} suggests to handle this by variable splitting, leading to the equivalent problem
\begin{eqnarray}\label{eqn:VarSplit1}
\left\{{\widehat \vx}_{MAP},{\widehat \vv} \right\} = \argmin{\vx, \vv} \ell(\vy , \vx) + \lambda \rho(\vv)~~~~s.t.~~~\vx=\vv.
\end{eqnarray}
The constraint is turned into a penalty term, relying on the augmented Lagrangian method (in its scaled dual form \cite{ADMM1}), leading to
\begin{eqnarray}\label{eqn:VarSplit2}
\left\{{\widehat \vx}_{MAP},{\widehat \vv} \right\} = \argmin{\vx, \vv} \ell(\vy , \vx) + \lambda \rho(\vv) + \frac{\beta}{2}\|\vx - \vv +\vu\|_2^2,
\end{eqnarray}
where $\vu$ serves as the Lagrange multiplier vector for the set of constraints. ADMM addresses the resulting problem by updating $\vx$, $\vv$, and $\vu$ sequentially in a block-coordinate-descent fashion, leading to the following series of sub-problems:
\begin{enumerate}
\item \textbf{Update of $\vx$:} When considering $\vv$ (and $\vu$) as fixed, the term $\rho(\vv)$ is omitted, and our task becomes
\begin{eqnarray}\label{eqn:UpdateX}
{\widehat \vx} = \argmin{\vx} \ell(\vy , \vx) + \frac{\beta}{2}\|\vx - \vv +\vu\|_2^2,
\end{eqnarray}
which is a far simpler inverse problem, where the regularization is an $L_2$ proximity one, which is easy to solve in most cases.

\item \textbf{Update of $\vv$:} In this stage we freeze $\vx$ (and $\vu$), and thus the log-likelihood term drops, leading to
\begin{eqnarray}\label{eqn:UpdateV}
{\widehat \vv} = \argmin{\vv}  \lambda \rho(\vv) + \frac{\beta}{2}\|\vx - \vv + \vu\|_2^2.
\end{eqnarray}
This stage is nothing but a denoising of the image $\vx+\vu$, assumed to be contaminated by a white additive Gaussian noise of power $\sigma^2 = 1/\beta$. This is easily verified by returning to Equation (\ref{eqn:BayesianMAP0}) and plugging the log-likelihood term $\|\vv-\vx-\vu\|_2^2/2\sigma^2$ referring to this case. Indeed, this is the prime observation in \cite{P3}, as they suggest to replace the direct solution of (\ref{eqn:UpdateV}) by activating an image denoising engine of choice. This way, we do not need to define explicitly the regularization $\rho(\cdot)$ to be used, as it is implied by the engine chosen.

\item \textbf{Update of $\vu$:} We complete the algorithm description by considering the update of the Lagrange multiplier vector $\vu$, which is done by ${\widehat \vu} = \vu + \vx -\vv$.

\end{enumerate}

\noindent
Although the above algorithm has a clear mathematical formulation and only two parameters, denoted by $ \beta $ and $ \lambda $, it turns out that tuning these is not a trivial task. The source of complexity emerges from the fact that the input noise-level to the denoiser is equal to $ \sqrt{\lambda/\beta} $. The confidence in the prior is determined by $ \lambda $, and the penalty on the distance between $ \vx $ and $ \vv $ is affected by $ \beta $. Empirically, setting a fixed value of $ \beta $ does not seize the potential of this algorithm; following previous work (e.g. \cite{Follows6,Follows11}), a common practical strategy to achieve a high-quality estimation is to increase the value of $ \beta $ as a function of the iterations: Starting from a relatively small value, i.e allowing an aggressive regularization, then proceeding to a more conservative one that limits the smoothing effect, up-to a point where $ \beta $ should be large enough to ensure convergence \cite{Follows11} and to avoid an undesired over-smoothed outcome. As one can imagine, it is cumbersome to choose the rate in which $ \beta $ should be increased, especially because the corrupted image $ \vx + \vu $ is a function of the Lagrange multiplier, which varies through the iterations as well.

In terms of convergence, the $P^3$ scheme has been shown to be well-behaved under some conditions on the denoising algorithm used. While the work reported in \cite{Follows1} requires the denoiser to be a symmetric smoothing and non-expansive filter, the later work in \cite{Follows11} relaxes this condition to much simpler boundedness of the denoising effect. However, both these  prove at best a convergence to a steady-state outcome, which is very far from the desired claim of getting to the global minimizer of the overall objective function. The work reported in \cite{Follows2} offers clear conditions for a global convergence of $P^3$, requiring the denoiser to be non-expansive, and emerging as the minimizer of a convex functional. A recently released paper extends the above by using a specific GMM-based denoiser, showing that these two conditions are met, thus guaranteeing global convergence of their ADMM scheme \cite{PPP-Convergence}.

Indeed, in that respect, a delicate matter with the $P^3$ approach is the fact that given a choice of a denoising engine, it does not necessarily refer to a specific choice of a prior $\rho(\cdot)$, as not every such engine could have a MAP-oriented interpretation.  This implies a fundamental difficulty in the $P^3$ scheme, as in this case we will be activating a denoising algorithm while departing from the original setting we have defined, and having no underlying cost function to serve. Indeed, the work reported in \cite{Follows1} addresses this very matter in a narrower setting, by studying the identity of the effective prior obtained from a chosen denoising engine. The author chooses to limit the answer to symmetric smoothing filters, showing that even in this special case, the outcome is far from being trivial. As we are about to see in the next section, this shortcoming can be overcome by adopting a different regularization strategy.

%%%%%%%%%%%%%%%%%%%%%%%%%%%%%%%%%%%%%%%%%%%%%%%%%%%%%%%%%%%%%%%%%%%%%%%%%%%%%%%%
%%%%%%%%%%%%%%%%%%%%%%%%%%%%%%%%%%%%%%%%%%%%%%%%%%%%%%%%%%%%%%%%%%%%%%%%%%%%%%%%

\section{The {\em Image Denoising Engine}}\label{sec:Engine}

Image denoising is a special case of the inverse problem posed in Equation (\ref{eqn:BayesianMAP0}), referring to the case $\vy = \vx+ \ve $, where $\ve$ is white Gaussian noise contamination of variance $\sigma^2$. In this case, the MAP problem becomes
\begin{eqnarray}\label{eqn:BayesianMAP1}
{\widehat \vx}_{Denoise} = \argmin{\vx} \frac{1}{2\sigma^2} \|\vy - \vx\|_2^2 + \lambda \rho(\vx).
\end{eqnarray}
The {\em image denoising engine}, which is the focal point of this work, is any candidate solver to the above problem, under a specific choice of a prior. In fact, in this work we choose to widen the definition of the image denoising engine to be any function $f:[0,255]^n \longrightarrow [0,255]^n$ that maps an image $\vy$ to another image $f(\vy)$ of the same size, and which aims to treat the denoising problem by the operation ${\widehat \vx}=f(\vy)$, be it MAP-based, MMSE-based, or any other approach.

Below, we accompany the definition of a denoiser with few basic conditions on the function $f$. Just before doing so, we make the following broad observation: Among the various degradations that inverse problems come to remedy, removal of noise is fundamentally different. Consider the set of all reasonable ``natural'' images living on a manifold ${\cal M}$. If we blur any given image or down-scale it, it is still likely to live in ${\cal M}$. However, if the image is contaminated by an additive noise, it pops out of the manifold along the normal to ${\cal M}$ with high probability. Denoising is therefore a fundamental mechanism for an orthogonal ``projection'' of an image back onto ${\cal M}$\footnote{In the context of optimization, a smaller class of the general denoising algorithms we define are characterized as ``proximal operators'' \cite{Proximal}. These operators are in fact direct generalizations of orthogonal projections.}. This may explain why denoising is such a central operation, which has been so heavily studied. In the context of this work, in any given step of our iterations, this projection would allow us to project the temporary result back onto ${\cal M}$, so as to increase chances of getting a good-quality restored version of our image.

\subsection{Conditions and Properties of $f(\vx)$}

We pose the following two necessary conditions on $f(\vx)$ that will facilitate our later derivations. Both these conditions rely on the differentiability\footnote{A discussion on this requirement and possible ways to relax it appear in appendix D.} of the denoiser $f(\vx$).

\begin{itemize}

\item \textbf{Condition 1: (Local) Homogeneity.} A denoiser applied to a positively scaled image $f(c\vx)$ should result in a scaled version of the original image. More specifically, for any scalar $c\geq0$ we must have $f(c\vx) = c f(\vx)$. In this work we shall relax this condition and demand its satisfaction for $|c-1|\le \epsilon$ for a very small $\epsilon$.

A direct implication of the above property refers to the behavior of the directional derivative of the denoiser $f(\vx)$ along the direction $\vx$. This derivative can be evaluated as
\begin{equation}
\label{eq:gradient}
\nabla_{\vx} f(\vx)\: \vx = \frac{f(\vx+\epsilon \vx)- f(\vx)}{\epsilon}
\end{equation}
for a very small $\epsilon$. Invoking the homogeneity condition this leads to
\begin{eqnarray}\nonumber
\nabla_{\vx} f(\vx)\: \vx & = &  \frac{(1+\epsilon) f(\vx)- f(\vx)}{\epsilon}  \\
& = & f(\vx).
\end{eqnarray}
Thus, the filter $f(\vx)$ can be written as\footnote{This result is sometimes known as Euler's homogeneous function theorem \cite{Euler}.}
\begin{equation}
\label{eq:Euler3}
f(\vx) = \nabla_{\vx} f(\vx)\: \vx.
\end{equation}

\item \textbf{Condition 2: Strong Passivity.} The Jacobian $\nabla_{\vx} f(\vx)$ of the denoising algorithm is stable, satisfying the condition
\begin{eqnarray}
\eta \left(\nabla_{\vx} f(\vx) \right) \le 1,
\end{eqnarray}
where $\eta(\mA)$ is the spectral radius of the matrix $\mA$. We interpret this condition as the restriction of the denoiser not to magnify the norm of an input image since
\begin{eqnarray}
\|f(\vx)\| = \|\nabla_{\vx} f(\vx) \vx\| \le \eta\left(\nabla_{\vx} f(\vx) \right) \cdot \|\vx\|\le \|\vx\|.
\end{eqnarray}
Here we have relied on the relation $f(\vx) = \nabla_{\vx} f(\vx) \vx$ that has been established in Equation (\ref{eq:Euler3}). Note that we have chosen this specific condition over the natural weaker alternative, $\|f(\vx)\| \le \|\vx\|$, since the strong condition implies the weak one.

\end{itemize}

\noindent A reformulation of the denoising engine that will be found useful throughout this work is the one suggested in \cite{Milanfar_SPM}, where we assume that the algorithm is built of two phases -- a first in which highly non-linear decisions are made, and a second in which these decisions are used to adapt a linear filter to the raw noisy image in order to perform the actual noise removal. Algorithms such as the NLM, kernel-regression, K-SVD, and many others admit this structure, and for them we can write
\begin{eqnarray}\label{eqn:linearfilter}
\widehat{\vx}_{\text{Denoise}} = f(\vy) = \bW (\vy) \vy.
\end{eqnarray}
The matrix $\bW$ is an $n\times n$ matrix, representing the (pseudo-) linear filter that multiplies the $n\times 1$ noisy image vector $\vy$. This matrix is image dependent, as it draws its content from the pixels in $\vy$. Nevertheless, this pseudo-linear format provides a convenient description of the denoising engine for our later derivations. We should emphasize that while this notation is true for only some of the denoising algorithms, the proposed framework we outline in this paper is general and applies to {\em any} denoising filter that satisfies the two conditions posed above. Indeed, the careful reader will observe that this pseudo-linear form is closely related to the directional derivative relation shown above: $f(\vy) = \nabla_{\vy} f(\vy)\: \vy$. In this form, the right hand side is now reminding us of the pseudo-linear form where the matrix $\mW(\vy)$ is replaced by the Jacobian matrix $ \nabla_{\vy} f(\vy)$.

As a side note we mention that yet another natural requirement on $\bW$ (or $\nabla_{\vy} f(\vy)$ in a wider perspective) is that it is row-stochastic, implying that (i) this matrix is (entry-wise) non-negative, and that (ii) the vector $\mathbb{1}$ is an eigenvector of $\bW$. This would imply a constancy behavior -- a denoiser $f(\vy)$ does not alter a constant image. More specifically, defining $\mathbb{1}$ as an $n$-dimensional column vector of all ones, for any scalar $c\geq0$ we have $f(c\mathbb{1}) = c\mathbb{1}$. Algorithms such as the NLM \cite{NLM1} and its many variants all lead to such a row-stochastic Jacobian. We note that this property, while nice to have, is not required for the derivations in this paper.

An interesting consequence of the homogeneity property is the following stability of the pseudo-linear operator $\bW$. Starting with a first-order Taylor expansion of $f(\vx+\vh)$, and invoking the directional derivative relation $f(\vy) = \nabla_{\vy} f(\vy)\: \vy$, we get
\begin{eqnarray}\label{eqn:stabilityW}
f(\vy + \vh) & \approx &  f(\vy) + \nabla_{\vy} f(\vy) \cdot \vh \\
& \approx & \nabla f(\vy) \cdot \vy + \nabla_{\vy} f(\vy) \cdot \vh \nonumber  \\
& \approx & \nabla_{\vy} f(\vy) \cdot (\vy + \vh). \nonumber
\end{eqnarray}
This result implies that while $\bW(\vy) = \nabla_{\vy} f(\vy)$ may indeed depend on $\vy$, its sensitivity to its perturbation is negligible, rendering it as an essentially constant linear operator on the perturbed image $\vy+\vh$.

\subsection{Denoisers Obeying the Above Conditions}

We cannot conclude this section without answering the key question: Which are the denoising engines to which we are constantly referring? While these could include any of the thousands of denoising algorithms published over the years, we obviously focus on the best performing ones, such as the Non-Local Means (NLM) and its advanced variants \cite{NLM1,NLM2,NLM3}, the K-SVD denoising method that relies on sparse representation modeling of image patches \cite{KSVDden} and its non-local extension \cite{Mairal}, the kernel-regression method that exploits local orientation \cite{KerReg}, the well-known BM3D that combines sparsity and self-similarity of patches \cite{BM3D}, the EPLL scheme that suggests patch-modeling based on the GMM model \cite{GMM1}, CSR and NCSR, which cluster the patches and sparsifies them jointly \cite{CSR,NCSR}, the group-Wiener filtering applied on patches \cite{ChattPeyman}, the multi-layer Perceptron method trained to clean an image or the more recent CNN-based alternative called Trainable Nonlinear Reaction-Diffusion (TNRD) algorithm \cite{NNden,chen2015trainable}, more recent work that proposed low-rank modeling of patches and the use of the weighted nuclear-norm \cite{WNNM}, non-local sparsity with GSM model \cite{NL-GSM}, and the list goes on and on. Each and every one of these options (and many others) is a candidate engine that could be fit into our scheme.

A fair and necessary question is whether the denoisers we work with obey the two conditions we have posed above (homogeneity and passivity), and whether the preliminary requirement of differentiability is met. We choose to defer the discussion on the differentiability to Appendix D, due to its relevance to several spread parts of this paper and focus here on the homogeneity and passivity.

\begin{figure*}
	\begin{center}
		\subfigure[K-SVD ($2.95e-4$)]{\includegraphics[width=0.3\linewidth]{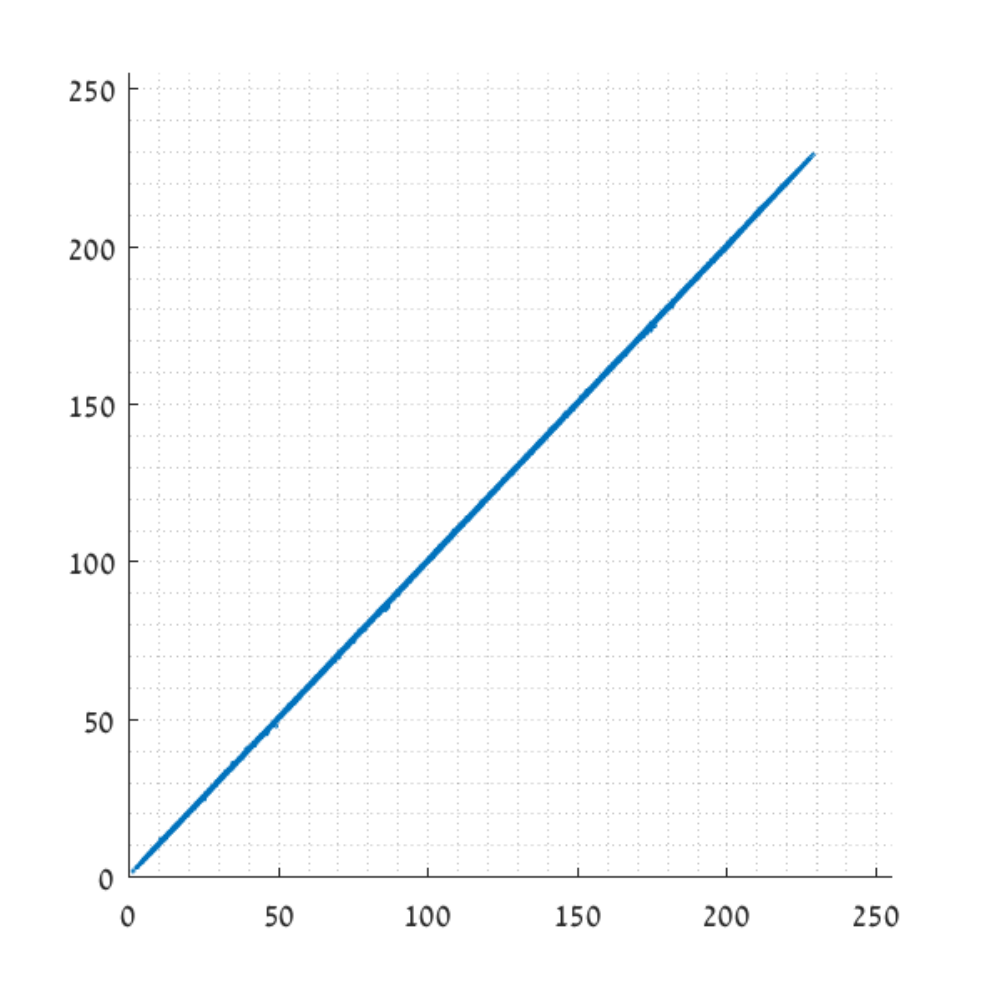}}
		\subfigure[BM3D ($3.38e-4$)]{\includegraphics[width=0.3\linewidth]{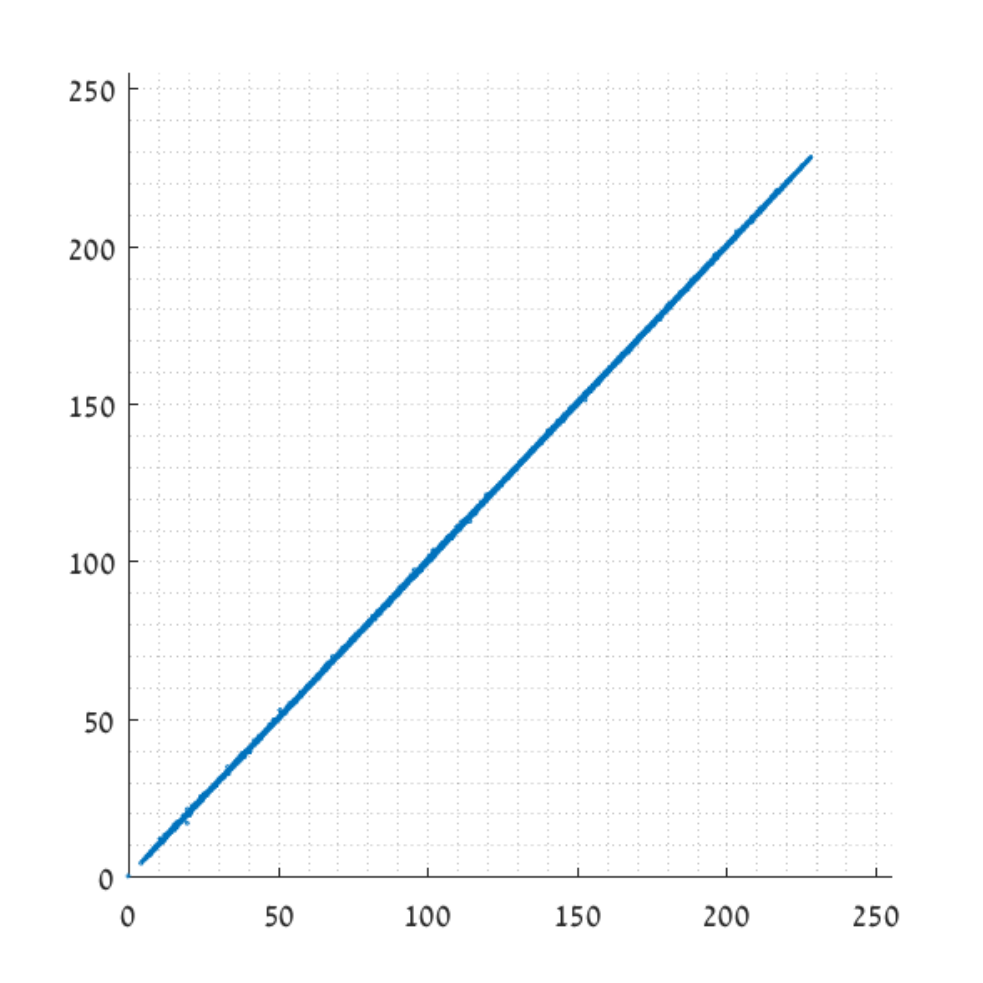}}
		\subfigure[NLM ($1.38e-4$)]{\includegraphics[width=0.3\linewidth]{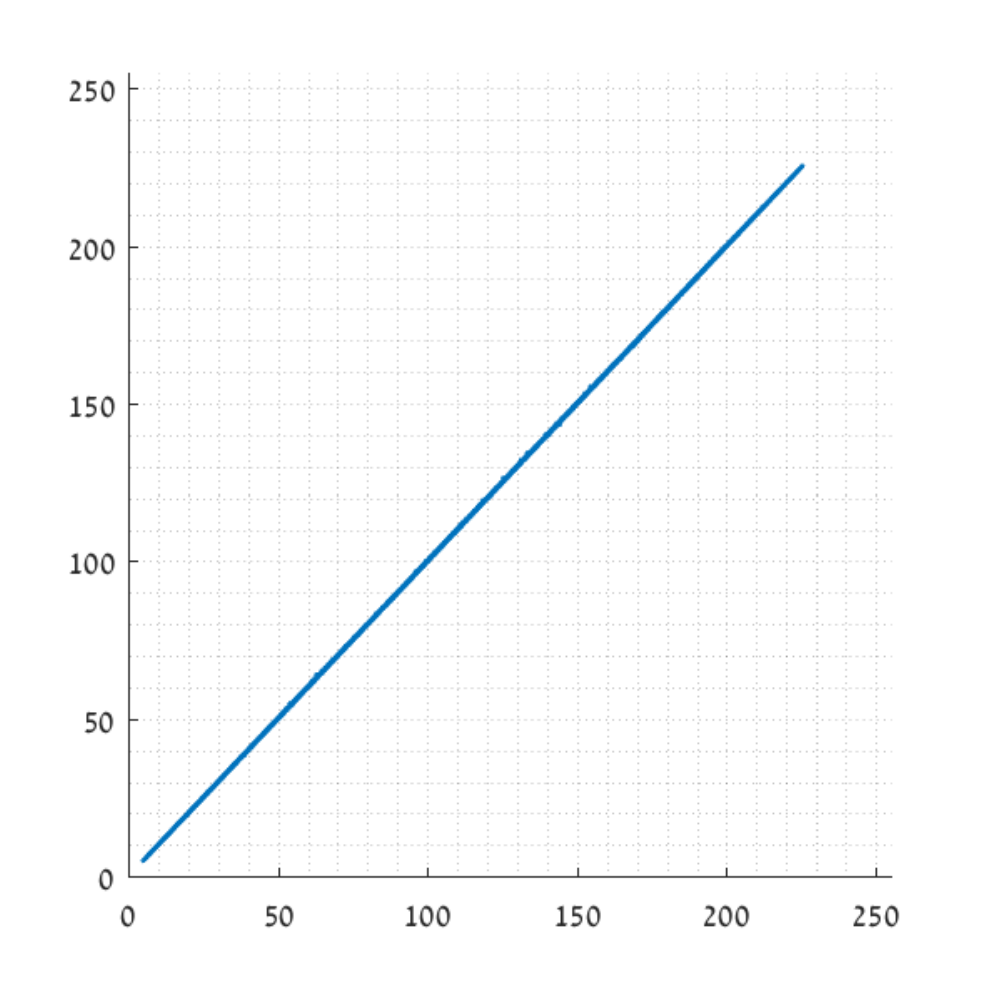}}\\
		\subfigure[EPLL ($1.46e-4$)]{\includegraphics[width=0.3\linewidth]{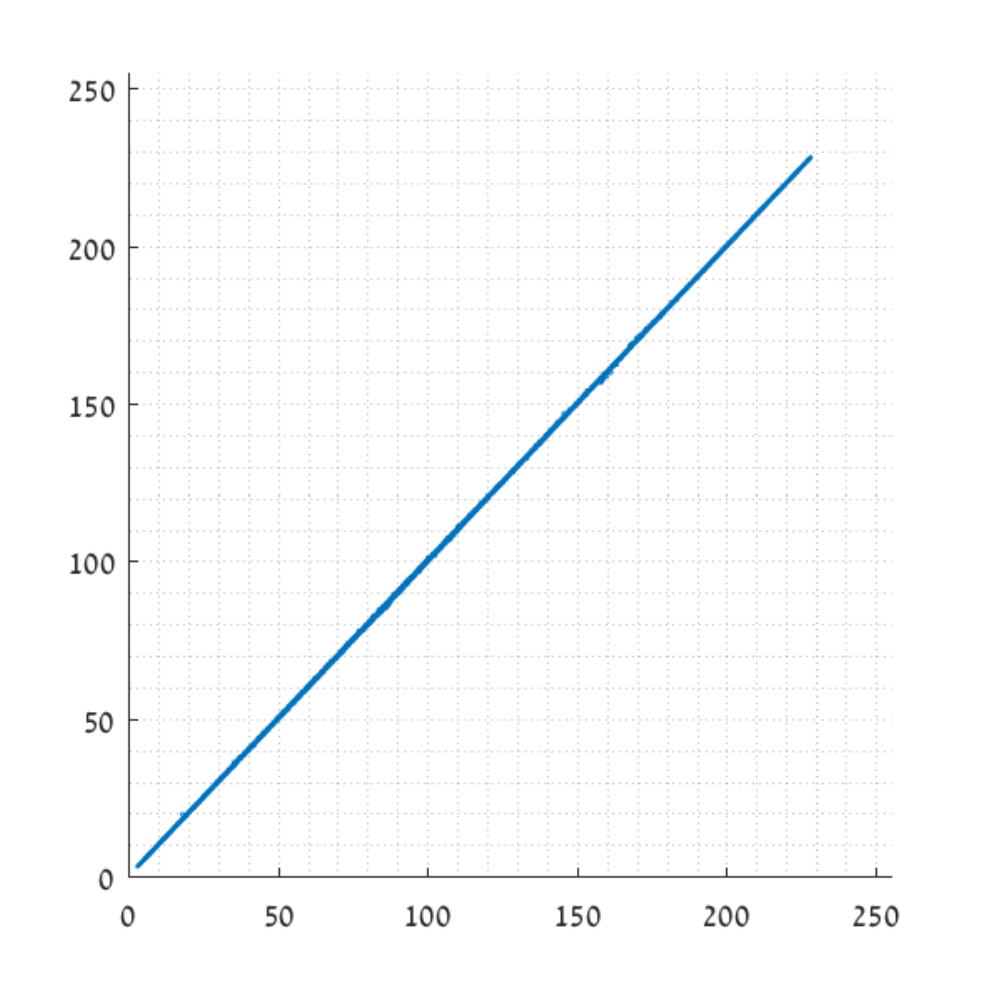}}
		\subfigure[TNRD ($9.51e-5$)]{\includegraphics[width=0.3\linewidth]{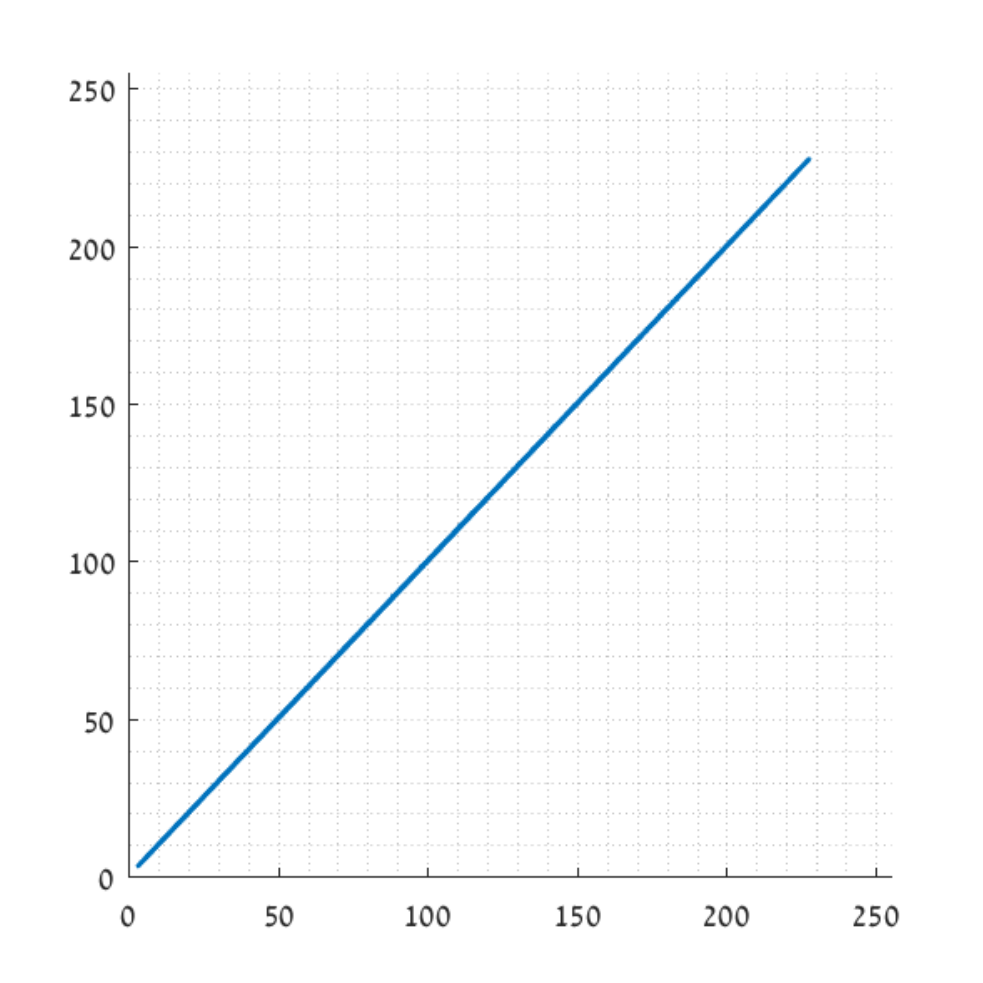}}
	\end{center}
	\caption{An empirical evaluation of the homogeneity property. These graphs show $f((1+\epsilon)\vx)$ versus $(1+\epsilon)f(\vx)$ as a scatter-plot for K-SVD, BM3D, NLM, EPLL, and the TNRD. Equality implies satisfaction of the homogeneity, and the numbers in the brackets provide the STD of the difference. {Note that these results were observed on various test images, but shown here for the image \textsf{Peppers}.} }
	\label{fig:Homogeneity-Demo}
\end{figure*}

Starting with the homogeneity property, we give an experimental evidence, accompanied by a theoretical analysis, to substantiate the fulfillment of this property by a series of well-known denoisers. Figure \ref{fig:Homogeneity-Demo} shows $f((1+\epsilon)\vx)$ versus $(1+\epsilon)f(\vx)$ as a scatter-plot, tested for K-SVD, BM3D, NLM, EPLL, and the TNRD. In all these experiments, the image $\vx$ is set to be \textsf{Peppers}, $\epsilon=0.01$ and $\sigma=5$ (level of noise assumed within $f$). As can be seen, a tendency to an equality $f((1+\epsilon)\vx) \approx (1+\epsilon)f(\vx)$ is obtained, suggesting that all these are indeed satisfying the homogeneity property. The deviation from exact equality in each of these tests has been evaluated as the standard deviation of the difference $f((1+\epsilon)\vx) - (1+\epsilon)f(\vx)$, leading to $2.95e-4,~3.38e-4,~1.38e-4,~1.46e-4,~9.51e-5$, respectively. A further discussion on the homogeneity property from a theoretical perspective is given in Appendix C.

Turning to the passivity condition, a conceptual difficulty is the need to explicitly obtain the Jacobian of the denoiser in question. Assuming that we overcame this problem somehow and got $\nabla_{\vx} f(\vx)$, its spectral radius would be evaluated using the Power-Method \cite{Power-Method} that applies iterations of the form
\begin{eqnarray}
\vh_{k+1} = \frac{\nabla_{\vx} f(\vx)\cdot \vh_{k}}{\| \nabla_{\vx} f(\vx) \cdot \vh_{k}\|_2}.
\end{eqnarray}
The spectral radius itself is easily obtained as
\begin{eqnarray}
\eta \left(\nabla_{\vx} f(\vx)\right) = \frac{\vh_{k+1}^T \vh_{k}}{\vh_{k}^T\vh_{k}}.
\end{eqnarray}
In order to bypass the need to explicitly obtain $\nabla_{\vx} f(\vx) \vh_{k}$, we rely on the first order Taylor expansion again,
\begin{eqnarray}
f(\vx+\vh) = f(\vx)+ \nabla_{\vx} f(\vx) \cdot \vh,
\end{eqnarray}
implying that $\nabla_{\vx} f(\vx) \cdot \vh \approx f(\vx+\vh) - f(\vx)$, which holds true if $\|\vh\|_2$ is small enough. Thus, our alternative Power-Method activates one denoising step per iteration,
\begin{eqnarray}
\vh_{k+1} \approx \frac{ f(\vx+\vh_k) - f(\vx)}{\| f(\vx+\vh_k) - f(\vx)\|_2}.
\end{eqnarray}
{The vector $\vh_k$ is normalized in each iteration, and thus $\|\vh_k\|_2=1$. This vector is an image, and thus the gray values in it must be very small ($\vh_k(j) \ll 1$), so as to lead to a sum of squares to be equal to 1. This agrees with the need for the perturbation $ \vx+\vh_k $ to be small.} 

%Since we normalize the vector $\vh_k$ after each iteration, the assumption that $\|\vh\|_2$ is small enough is met, as its gray-values are much smaller than $1$.

This algorithm has been applied to K-SVD, BM3D, NLM, EPLL, and the TNRD ($\vx$ set to be the image \textsf{Cameraman}, $\sigma=5$, number of iterations set to give an accuracy of $1e-5$), resulting all with values smaller or equal to $1$, verifying the passivity of these filters.

%%%%%%%%%%%%%%%%%%%%%%%%%%%%%%%%%%%%%%%%%%%%%%%%%%%%%%%%%%%%%%%%%%%%%%%%%%%%%%%%
%%%%%%%%%%%%%%%%%%%%%%%%%%%%%%%%%%%%%%%%%%%%%%%%%%%%%%%%%%%%%%%%%%%%%%%%%%%%%%%%

\section{Regularization by Denoising (RED)}\label{sec:Algorithm}

%%%%%%%%%%%%%%%%%%%%%%%%%%%%%%%%%%%%%%%%%%%%%%%%%%%%%%%%%%%%%%%%%%%%%%%%%%%%%%%%
%%%%%%%%%%%%%%%%%%%%%%%%%%%%%%%%%%%%%%%%%%%%%%%%%%%%%%%%%%%%%%%%%%%%%%%%%%%%%%%%

\subsection{The Image-Adaptive Laplacian}

The new and alternative framework we propose relies on a form of an image-adaptive Laplacian which builds a powerful (empirical) prior that can be used to regularize a variety of inverse problems. As a place to start and motivate this definition, let's go back to the description of the denoiser given in Equation (\ref{eqn:linearfilter}), namely\footnote{Note that we conveniently assume that the prior is applied to the clean image $\vx$, a matter that will be clarified as we dive into our explanations.} $\mW(\vx) \vx$. We may think of this pseudo-linear filter as one where a set of coefficients (depending on $\vx$) are first computed in the matrix $\bW$, and then applied to the image $\vx$. From this we can construct the Laplacian form,
\begin{equation}
\rho_L(\vx) = \frac{1}{2} \: \vx^T \mL(\vx) \vx = \frac{1}{2} \: \vx^T (\mI - \mW(\vx)) \vx = \frac{1}{2} \: \vx^T \left[\vx- \mW (\vx) \vx \right].
\end{equation}
This definition by itself is not novel, as it is similar to ideas brought up in a series of recent contributions
\cite{LapReg1,LapReg2,LapReg3,LapReg4,Milanfar_SPM,LapReg5,LapReg6,LapReg7,LapReg8,LapReg9}. This expression relies on using an image-adaptive Laplacian -- one that draws its definition from the image itself.

Observing the obtained expression, we note that it can be interpreted as the {\em unnormalized cross-correlation} between the image $\vx$ and its corresponding residual $\vx-\mW(\vx) \vx $. As a {\em prior} expression should give low values for likely images, in our case this would be achieved in one of two ways (or their combination):
\begin{itemize}
\item A small value is obtained for $\rho_L(\vx)$ if the residual is very small, implying that the image $\vx$ serves as a near fixed-point of the denoising engine, $\vx \approx \mW(\vx) \vx$.
\item A small value is obtained for $\rho_L(\vx)$  if the cross-correlation of the residual to the image itself is small, a feature that implies that the residual behaves like white noise, or alternatively, if it does not contain elements from the image itself. Interestingly, this concept has been harnessed successfully by some denoising algorithms such as the Dantzig-Selector \cite{Dantzig} and by image denoising boosting techniques \cite{Milanfar_SPM,Boosting1,Boosting2}. Indeed, enforcing orthogonality between the signal and its treated residual is the underlying force behind the Normal equations in statistical estimation (e.g. Least Squares and Kalman Filtering).% \cite{Estimation}.
\end{itemize}

\noindent Given the above prior, we return to the general inverse-problem posed in Equation (\ref{eqn:BayesianMAP0}), and define our new objective,
\begin{eqnarray}\label{eqn:BayesianMAPLap1}
{\widehat \vx} = \argmin{\vx} \ell(\vy , \vx) + \frac{\lambda}{2} \: \vx^T \left[\vx- \mW (\vx) \vx\right].
\end{eqnarray}
The prior expression, while exhibiting a possibly complicated dependency on the unknown $\vx$, is well-defined and clear. Nevertheless, an attempt to apply any gradient-based algorithm for solving the above minimization task encounters an immediate problem, due to the need to differentiate $\bW(\vx)$ with respect to $\vx$. We overcome this problem by observing that $\bW(\vx) \vx$ is in fact the activation of the image denoising engine on $\vx$, i.e., $f(\vx) = \bW(\vx) \vx$. This observation inspires the following more general definition of the Laplacian regularizer, which is the prime message of this paper:
\begin{equation}
\rho_L(\vx) = \frac{1}{2} \: \vx^T\left[\vx-f(\vx)\right].
\end{equation}
This is the Regularization by Denoising (RED) paradigm that this work advocates. In this expression, the residual is defined more generally for any filter $f(\vx)$ even if it can not be written in the familiar (pseudo-)linear form. Note that all the preceding intuition about the meaning of  this prior remains intact; namely, the value is low if the cross-correlation between the image and its denoising residual is small, or if the residual itself is small due to $\vx$ being a fixed point of $f$.

Surprisingly, while this expression is more general, it leads to a better-managed optimization problem due to the careful properties we have outlined in Section \ref{sec:Engine} on our denoising engines $f$. The overall energy functional to minimize is
\begin{eqnarray}\label{eqn:BayesianMAPLap2}
E(\vx) = \ell(\vy , \vx) + \frac{\lambda}{2} \: \vx^T\left(\vx- f(\vx)\right),
\end{eqnarray}
and the gradient of this expression is readily available by
\begin{eqnarray}\label{eqn:Gradient1}
\nabla_{\vx} E(\vx)  & = & \nabla_{\vx} \ell(\vy , \vx) + \frac{\lambda}{2} \: \nabla_{\vx} \left\{ \vx^T\left(\vx-f(\vx)\right)\right\} \\ \nonumber
& = & \nabla_{\vx} \ell(\vy , \vx) + \frac{\lambda}{2} \: \nabla_{\vx} \vx^T\vx - \frac{\lambda}{2} \: \nabla_{\vx} \left[\vx^T f(\vx) \right]\\ \nonumber
& = & \gradient_{\vx} \ell(\vy , \vx) + \lambda \vx - \frac{\lambda}{2}\left[ f(\vx) + \nabla_{\vx} f(\vx) \vx \right].
\end{eqnarray}
Based on our prior assumption regarding the availability of a directional derivative for the denoising engine, the term $\nabla_{\vx} f(\vx) \vx$ can be replaced\footnote{A better approximation can be applied in which we replace $\nabla_{\vx} f(\vx) \vx$ by the difference $(f((1+\epsilon)\vx)-f(\vx))/\epsilon$, but this calls for two activations of the denoising engine per gradient evaluation.} by
$\nabla_{\vx} f(\vx)\: \vx =  f(\vx)$, based on Equation (\ref{eq:Euler3}), implying that the gradient expression is further simplified to be
\begin{eqnarray}\label{eqn:Gradient3}
\nabla_{\vx} E(\vx)  = \gradient_{\vx} \ell(\vy , \vx) + \lambda \vx - {\lambda} f(\vx) = \gradient_{\vx} \ell(\vy , \vx) + \lambda (\vx - f(\vx)),
\end{eqnarray}
requiring only one activation of the denoising engine for the gradient evaluation. Interestingly, if we bring back now the pseudo-linear interpretation of the denoising engine, the gradient would be the residual, just as posed above, implying that
\begin{eqnarray}\label{eqn:Gradient4}
\nabla_{\vx} E(\vx)  = \gradient_{\vx} \ell(\vy , \vx) + \lambda (\vx - \bW(\vx) \vx).
\end{eqnarray}
Observe that this is a non-trivial derivation of the gradient of the original penalty function posed in Equation (\ref{eqn:BayesianMAPLap1}).

%%%%%%%%%%%%%%%%%%%%%%%%%%%%%%%%%%%%%%%%%%%%%%%%%%%%%%%%%%%%%%%%%%%%%%%%%%%%%%%%
%%%%%%%%%%%%%%%%%%%%%%%%%%%%%%%%%%%%%%%%%%%%%%%%%%%%%%%%%%%%%%%%%%%%%%%%%%%%%%%%

\subsection{Deploying the Denoising Engine for Solving Inverse Problems}

\begin{figure}[hbtp]
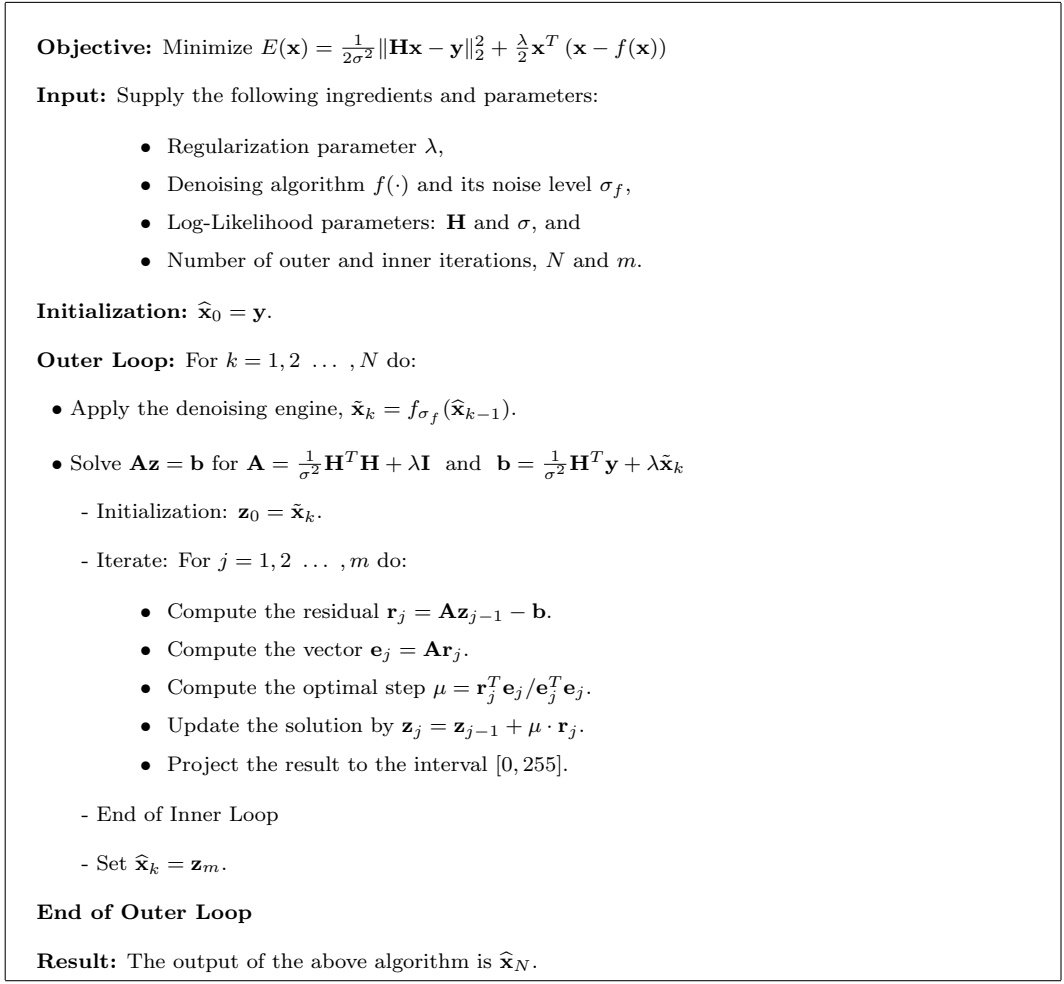

\begin{center}
\begin{tabular}{|c|}
\hline
\begin{minipage}[b]{0.8\linewidth}
    \renewcommand{\baselinestretch}{1}
    \small \vspace{0.1in}
\scriptsize{
    \begin{description}
    \item {\bf Objective:} Minimize $E(\vx) = \frac{1}{2\sigma^2}\|\mH\vx -\vy\|_2^2 + \frac{\lambda}{2} \vx^T\left(\vx-f(\vx)\right)$
    \item {\bf Input:} Supply the following ingredients and parameters:
    \begin{itemize}
            \item Regularization parameter $\lambda$,
            \item Denoising algorithm $f(\cdot)$ and its noise level $\sigma_f$,
            \item Log-Likelihood parameters: $\mH$ and $\sigma$, and
            \item Number of outer and inner iterations, $N$ and $m$.
    \end{itemize}
    \item {\bf Initialization:} ${\widehat \vx}_0 = \vy$.
    \item {\bf Outer Loop:} For $k=1,2 ~\ldots~, N$ do:
    \item ~~$\bullet$ Apply the denoising engine, ${\tilde \vx}_k=f_{\sigma_f}({\widehat \vx}_{k-1})$.
    \item ~~$\bullet$ Solve $\mA \vz = \vbee$ for $\mA = \frac{1}{\sigma^2} \mH^T \mH + \lambda \mI~~\mbox{and}~~\vbee= \frac{1}{\sigma^2}\mH^T \vy + \lambda {\tilde \vx}_k$
    \item ~~~~~~- Initialization: ${\vz}_0 = {\tilde \vx}_k$.
    \item ~~~~~~- Iterate: For $j=1,2 ~\ldots~, m$ do:
               \begin{itemize}
                  \item Compute the residual $\vr_j = \mA \vz_{j-1} - \vbee$.
                  \item Compute the vector $\ve_j = \mA \vr_j$.
                  \item Compute the optimal step $\mu = \vr_j^T \ve_j/\ve_j^T \ve_j$.
                  \item Update the solution by $\vz_j=\vz_{j-1}+\mu\cdot \vr_j$.
                  \item Project the result to the interval $[0,255]$.
               \end{itemize}
    \item ~~~~~~- End of Inner Loop
    \item ~~~~~~-  Set ${\widehat \vx}_k = \vz_m$.
    \item {\bf End of Outer Loop}
    \item {\bf Result:} The output of the above algorithm is ${\widehat \vx}_N$.

    \end{description}
}
\end{minipage}
\\ \hline
\end{tabular}
\\ \vspace{0.1in}
\caption{The Proposed Scheme (RED) via the fixed-point method.} \label{fig:algoFP}
\end{center}
\end{figure}

In the discussion above we have seen that the gradient of the energy functional to minimize (given in Equation (\ref{eqn:BayesianMAPLap2})) is easily computable, given in Equation (\ref{eqn:Gradient3}). We now turn to show several options for using this in order to solve a general inverse problem. Common to all these methods is the fact that the eventual algorithm is iterative, in which each step is composed of applying the denoising engine (once or more), accompanied by other simpler calculations. In Figures \ref{fig:algoSD}, \ref{fig:algoADMM}, and \ref{fig:algoFP} we present pseudo-code for several such algorithms, all in the context of handling the case in which the likelihood function is given by $\ell(\vy,\vx) = \|\mH\vx - \vy\|_2^2/2\sigma^2$.

\begin{itemize}

\item \textbf{Gradient Descent Methods:} Given the gradient of the energy function $E(\vx)$, the Steepest-Descent (SD) is the simplest option that can be considered, and it amounts to the update formula
\begin{eqnarray}\label{eqn:SD}
{\widehat \vx}_{k+1} & = & {\widehat \vx}_{k} - \mu \nabla_{\vx} \left. E(\vx)\right|_{{\widehat \vx}_{k}} \\ \nonumber & = & {\widehat \vx}_{k} - \mu \left[\gradient_{\vx} \left. \ell(\vy , \vx)\right|_{{\widehat \vx}_{k}} + \lambda ({\widehat \vx}_{k} -  f({\widehat \vx}_{k}))\right].
\end{eqnarray}
Figure \ref{fig:algoSD} describes this algorithm in more details.
% {\color{blue} WE NEED TO JUSTIFY THE FORMULA $\mu = 2/(1/\sigma^2 +\lambda)$ THAT WE SUGGEST IN THE ALGORITHM DESCRIPTION.}

A line-search can be proposed in order to set $\mu$ dynamically per iteration, but this is necessarily more involved. For example, in the case of the Armijo rule, it requires a computation of the above gradient $\vgee_k$ and then assessing the energy $E( {\widehat \vx}_{k} - \mu \vgee_k)$ for different values of $\mu$ in a retracting fashion, each of which calling for a computation of the denoising engine once.

One could envision using the Conjugate-Gradient (CG) to speed this method, or better yet, applying the Sequential Subspace Optimization (SESOP) algorithm \cite{SESOP}. SESOP holds the current gradient and the last several update directions as the columns of a matrix $\mV_k$ (referring to the $k^{th}$ iteration), and seeks the best linear combination of these columns as an update direction to the current solution, namely $\vx_{k+1} = \vx_k + \mV_k \va_k$. When restricted to have only one column, this reduces to a simple SD with line-search. When using two columns, it has the flavor (and strength) of CG, and when using more columns, this method can lead to much faster convergence in non-quadratic problems. The key points of SESOP are (i) The matrix $\mV$ is updated easily from one iteration to another by discarding the last direction, bringing in the last one, and adding the new gradient; and  (ii) The unknown weights vector $\va_k$ is low-dimensional, and thus updating it can be done using a Newton method. Naturally, one should evaluate the first and second derivatives of the penalty function w.r.t. $\va_k$, and these will leverage the relations established above. We shall not dive deeper into this option because it will not be included in our experiments.

One possible shortcoming of the gradient approach (in all its manifestations) is the fact that per activation of the denoising engine, the likelihood is updated rather mildly as a simple step toward the current log-likelihood gradient. This may imply that the overall algorithm will require many iterations to converge. The next two methods propose a way to overcome this limitation, by treating the log-likelihood term more ``aggressively''.

\begin{figure}[hbtp]
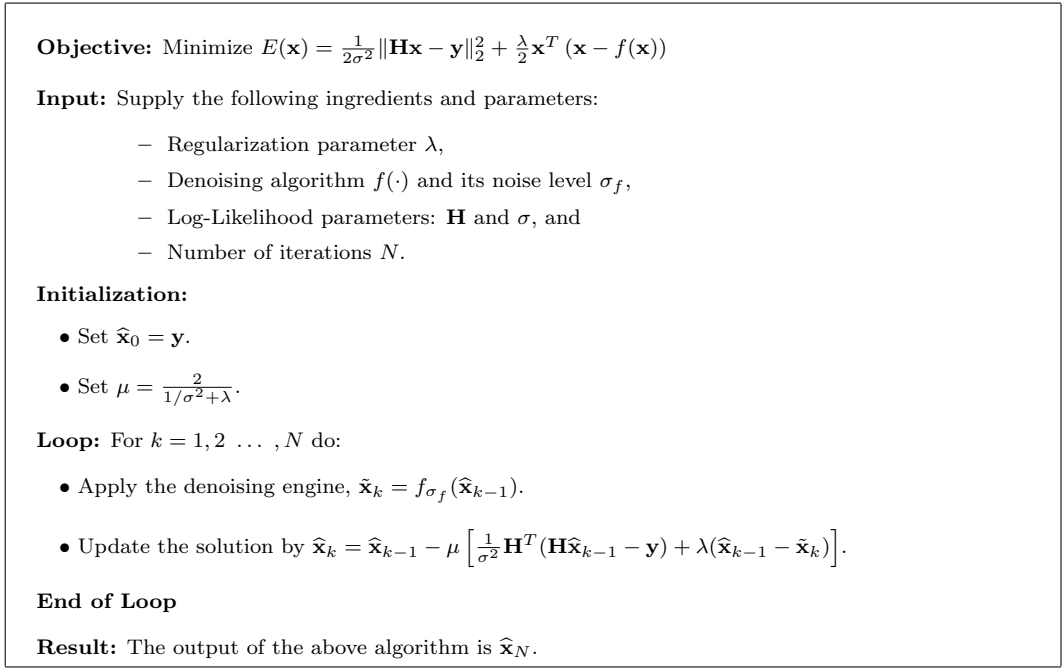

\begin{center}
\begin{tabular}{|c|}
\hline
\begin{minipage}[b]{0.8\linewidth}
    \renewcommand{\baselinestretch}{0.5}
    \small \vspace{0.1in}
\scriptsize{
    \begin{description}
    \item {\bf Objective:} Minimize $E(\vx) = \frac{1}{2\sigma^2}\|\mH\vx -\vy\|_2^2 + \frac{\lambda}{2} \vx^T\left(\vx-f(\vx)\right)$
    \item {\bf Input:} Supply the following ingredients and parameters:
    \begin{itemize}
            \item Regularization parameter $\lambda$,
            \item Denoising algorithm $f(\cdot)$ and its noise level $\sigma_f$,
            \item Log-Likelihood parameters: $\mH$ and $\sigma$, and
            \item Number of iterations $N$.
    \end{itemize}
    \item {\bf Initialization:}
    \item ~~~$\bullet$ Set ${\widehat \vx}_0 = \vy$.
    \item ~~~$\bullet$ Set $\mu = \frac{2}{1/\sigma^2 +\lambda}$.
    \item {\bf Loop:} For $k=1,2 ~\ldots~, N$ do:
    \item ~~~$\bullet$ Apply the denoising engine, ${\tilde \vx}_k=f_{\sigma_f}({\widehat \vx}_{k-1})$.
    \item ~~~$\bullet$ Update the solution by ${\widehat \vx}_k = {\widehat \vx}_{k-1} - \mu\left[\frac{1}{\sigma^2}\mH^T (\mH {\widehat \vx}_{k-1}- \vy) +\lambda ({\widehat \vx}_{k-1}-{\tilde \vx}_{k})\right]$.
    \item {\bf End of Loop}
    \item {\bf Result:} The output of the above algorithm is ${\widehat \vx}_N$.

    \end{description}
}
\end{minipage}
\\ \hline
\end{tabular}
\\ \vspace{0.1in}
\caption{The proposed scheme (RED) via the steepest descent method.} \label{fig:algoSD}
\end{center}
\end{figure}

\item \textbf{ADMM:} Addressing the optimization task given in Equation  (\ref{eqn:BayesianMAPLap2}), we can imitate the path taken by the $P^3$ scheme, and apply variable splitting and ADMM. The steps would follow the description given in Section \ref{sec:Background} almost exactly, with one major difference -- the prior in our case is explicit, and therefore, the stage termed ``update of $\vv$'' would become
\begin{eqnarray}\label{eqn:UpdateVours}
{\widehat \vv} = \argmin{\vv}  \frac{\lambda}{2} \vv^T\left(\vv-f(\vv)\right) + \frac{\beta}{2}\|\vv - \vx - \vu\|_2^2.
\end{eqnarray}
Rather than applying an arbitrary denoising engine to compute ${\widehat \vv}$ as a replacement to the actual minimization, we should target this minimization directly by some iterative scheme. For example, setting the gradient of the above expression to zero leads to the equation
\begin{eqnarray}\label{eqn:UpdateVours1}
\lambda (\vv-f(\vv)) + \beta(\vv - \vx - \vu)= \vzero,
\end{eqnarray}
which can be solved iteratively using the fixed-point strategy, by
\begin{eqnarray}\label{eqn:UpdateVours2}
\lambda (\vv_j-f(\vv_{j-1})) + \beta(\vv_{j} - \vx - \vu)= \vzero
\\ \nonumber \rightarrow~~
\vv_j = \frac{1}{\beta+\lambda}\left(  \lambda f(\vv_{j-1} + \beta (\vx + \vu) \right).
\end{eqnarray}
This means that our approach in this case is computationally more expensive, as it will require several activations of the denoising engine. {However, a common approach to speed up the convergence (in terms of runtime) of the ADMM is called ``early termination'' \cite{ADMM1}, suggesting to approximate the solution of the $ \vv $-update stage. We found this approach useful for our setting, especially because the application of a denoiser is computationally expensive. To this end, we may choose to apply only one iteration of the iterative process described in Equation \eqref{eqn:UpdateVours2}, which amounts to one operation of a denoising algorithm.} Figure \ref{fig:algoADMM} describes this specific algorithm in more details. If one changes all {\bf Part 2} (in Figure \ref{fig:algoADMM}) with the computation ${\widehat \vv}_k = f_{1/\sqrt{\beta}}(\vz^*)$, we obtain the $P^3$ scheme for the same choice of the denoising engine. While this difference is quite delicate, we should remind the reader that (i) this bridge between the two approaches is valid only when we deploy ADMM on our scheme, and (ii) as opposed to the $P^3$ method, our method is guaranteed to converge to the global optimum of the overall penalty function, as will be described hereafter.

We should point out that when using the ADMM, the update of $\vx$ applies an aggressive inversion of the log-likelihood term, which is followed by the above optimization task. Thus, the shortcoming mentioned above regarding the lack of balance between the treatments given to the likelihood and the prior is mitigated.

\begin{figure}[hbtp]
\begin{center}
\renewcommand{\baselinestretch}{1}
\begin{tabular}{|c|}
\hline
\begin{minipage}[b]{0.8\linewidth}
    \small \vspace{0.1in}
\scriptsize{
    \begin{description}
    \item {\bf Objective:} Minimize $E(\vx) = \frac{1}{2\sigma^2}\|\mH\vx -\vy\|_2^2 + \frac{\lambda}{2} \vx^T\left(\vx-f(\vx)\right)$
    \item {\bf Input:} Supply the following ingredients and parameters:
    \begin{itemize}
            \item Regularization parameter $\lambda$,
            \item Denoising algorithm $f(\cdot)$ and its noise level $\sigma_f$,
            \item Log-Likelihood parameters: $\mH$ and $\sigma$,
            \item Number of outer and inner iterations, $N$, $m_1$, and $m_2$, and
            \item ADMM coefficient $\beta$.
    \end{itemize}
    \item {\bf Initialization:} Set ${\widehat \vx}_0 = \vy$, ${\widehat \vv}_0 = \vy$, and ${\widehat \vu}_0 = \vzero$.
    \item {\bf Outer Loop:} For $k=1,2 ~\ldots~, N$ do:

    \item ~~~~~~ \textbf{Part 1:} Solve ${\widehat \vx}_k = \argmin{\vz} \frac{1}{2\sigma^2}\|\mH\vz - \vy\|_2^2+ \frac{\beta}{2}\|\vz-{\widehat \vv}_{k-1}+{\widehat \vu}_{k-1} \|_2^2$ by

        \item ~~~~~~~~~$\bullet$ Initialization: ${\vz}_0 = {\widehat \vx}_{k-1}$, and define $\vz^*={\widehat \vv}_{k-1}-{\widehat \vu}_{k-1}$.
        \item ~~~~~~~~~$\bullet$ Inner Iteration: For $j=1,2 ~\ldots~, m_1$ do:
        \begin{description}
            \item ~~~~$-$ Compute the gradient $\ve_j = \frac{1}{\sigma^2}\mH^T(\mH\vz_{j-1}-\vy)+\beta(\vz_{j-1}-\vz^*)$.
            \item ~~~~$-$ Compute $\vr_j= \frac{1}{\sigma^2}\mH^T\mH\ve_j+\beta\ve_j$.
            \item ~~~~$-$ Compute the step size $\mu= \ve_j^T \ve_j/\ve_j^T \vr_j$.
            \item ~~~~$-$ Update the solution by $\vz_j=\vz_{j-1}+\mu \ve_j$.
            \item ~~~~$-$ Project the result to the interval $[0,255]$.
        \end{description}
        \item ~~~~~~~~~$\bullet$ End of Inner Loop
        \item ~~~~~~~~~$\bullet$ Set ${\widehat \vx}_k = \vz_{m_1}$.

    \item ~~~~~~ \textbf{Part 2:} Solve  ${\widehat \vv}_k = \argmin{\vz} \lambda \vz^T (\vz-f_{\sigma_f}(\vz))+ \frac{\beta}{2}\|\vz-{\widehat \vx}_{k}-{\widehat \vu}_{k-1} \|_2^2$ by

        \item ~~~~~~~~~$\bullet$ Initialization: ${\vz}_0 = {\widehat \vv}_{k-1}$, and define $\vz^*={\widehat \vx}_{k}+{\widehat \vu}_{k-1}$.
        \item ~~~~~~~~~$\bullet$ Inner Iteration: For $j=1,2 ~\ldots~, m_2$ do:
            \begin{description}
                \item ~~~~$-$ Apply the denoising engine, ${\tilde \vz}_j=f_{\sigma_f}({\widehat \vz}_{j-1})$.
                \item ~~~~$-$ Compute the gradient $\vz_j = \frac{1}{\beta+\lambda} \left(\lambda {\tilde \vz}_j +\beta \vz^*\right)$.
            \end{description}
        \item ~~~~~~~~~$\bullet$ End of Inner Loop
        \item ~~~~~~~~~$\bullet$ Set ${\widehat \vv}_k = \vz_{m_2}$.

    \item ~~~~~~ \textbf{Part 3:} Update  ${\widehat \vu}_k = {\widehat \vu}_{k-1}+ {\widehat \vx}_{k}-{\widehat \vu}_{k}$.

    \item {\bf End of Outer Loop}
    \item {\bf Result:} The output of the above algorithm is ${\widehat \vx}_N$.

    \end{description}
}
\end{minipage}
\\ \hline
\end{tabular}
\\ \vspace{0.1in}
\caption{The proposed scheme (RED) via the ADMM method.} \label{fig:algoADMM}
\end{center}
\end{figure}

\item \textbf{Fixed-Point Strategy:} An appealing alternative to the above exists, obtained via the fixed-point strategy. As our aim is to find $\vx$ that nulls the gradient, this could be posed as an implicit equation to be solved directly,
\begin{eqnarray}
\gradient_{\vx} \ell(\vy , \vx) + \lambda (\vx -  f(\vx))=0.
\end{eqnarray}
Using the fixed-point strategy, this could be handled by the iterative formula
\begin{eqnarray}
\gradient_{\vx} \ell(\vy , \vx_{k+1}) + \lambda (\vx_{k+1} -  f(\vx_k))=0.
\end{eqnarray}
As an example, in the case of linear degradation model and Gaussian white additive noise, this equation would be
\begin{eqnarray}
\frac{1}{\sigma^2} \mH^T (\vy - \mH\vx_{k+1}) + \lambda (\vx_{k+1} -  f(\vx_k))=0,
\end{eqnarray}
leading to the recursive update relation
\begin{eqnarray}
\vx_{k+1}= \left[\frac{1}{\sigma^2} \mH^T \mH + \lambda \mI\right]^{-1} \left[ \frac{1}{\sigma^2}\mH^T \vy + \lambda f(\vx_k)\right].
\end{eqnarray}
This formula suggests one activation of the denoising per iteration, followed by what seems to be a plain Wiener filtering computation\footnote{Note that Equation (\ref{eqn:UpdateVours2}) refers to the same problem posed here under the choice $\mH=\mI$ and $\beta=1/\sigma^2$.}. The matrix inversion itself could be done in the Fourier domain for block-circulant $\mH$, or iteratively using for example, the Richardson algorithm: Defining
\begin{eqnarray}
\mA = \frac{1}{\sigma^2} \mH^T \mH + \lambda \mI~~~\mbox{and}~~~
\vbee= \frac{1}{\sigma^2}\mH^T \vy + \lambda f(\vx_k),
\end{eqnarray}
our goal is to solve the linear system $\mA\vx = \vbee$. This is achieved by a variant of the SD method\footnote{All this refer to a specific iteration $k$ within which we apply inner iterations to solve the linear system, and thus the use of the different index $j$.}, $\vx_{j+1} = \vx_j - \mu (\mA \vx_{j} - \vbee) = \vx_j - \mu \ve_j$, where we have defined $\ve_j = \mA \vx_{j} - \vbee$. By setting the step size to be $\mu_j=\ve_j^T \mA \ve_j/\ve_j^T \mA^T \mA \ve_j$, we greedily optimize the potential of each iteration.

Convergence of the above algorithm is guaranteed since
\begin{eqnarray}
\left\| \left[\frac{1}{\sigma^2} \mH^T \mH + \lambda \mI\right]^{-1} \lambda \nabla_{\vx} f(\vx_k)\right\|<1.
\end{eqnarray}
% {\color{blue} PROVE THIS}

This approach, similarly to the ADMM, has the desired balance mentioned above between the likelihood and the regularization terms, matching the efforts dedicated to both. A pseudo-code describing this algorithm appears in Figure \ref{fig:algoFP}.

\end{itemize}

\noindent A basic question that has not been discussed so far is how to set the parameters of $f(\vx)$ in defining the regularization term. More specifically, assuming that the denoising engine depends on one parameter -- the noise standard-deviation $\sigma_f$ -- the question is which value to use. While one could envision using varying values as the iterations progress and the outcome improves, the approach we take in this work is to set this parameter to be a small and fixed value. Our intuition for this choice is the desire to have a clear and fixed regularization term, which in turn implies a clear cost function to work with. Furthermore, the prior we propose should encapsulate in it our desire to get to a final image that is a stable point of such a weak denoising engine, $\vx \approx f(\vx)$. Clearly, more work is required to better understand the influence of this parameter and its automatic setting.

%%%%%%%%%%%%%%%%%%%%%%%%%%%%%%%%%%%%%%%%%%%%%%%%%%%%%%%%%%%%%%%%%%%%%%%%%%%%%%%
%%%%%%%%%%%%%%%%%%%%%%%%%%%%%%%%%%%%%%%%%%%%%%%%%%%%%%%%%%%%%%%%%%%%%%%%%%%%%%%%

\section{Analysis}\label{sec:Analysis}

\subsection{Convexity}

Is our proposed regularization function $\rho_L(\vx)$ convex? At first glance, this may seem like too much to expect. Nevertheless, it appears that for reasonably performing denoising engines obeying the conditions posed in Section \ref{sec:Engine}, this is exactly the case. For the function $\rho_L(\vx) = \vx^T(\vx-f(\vx))$ to be convex, we should demand that the second derivative is a positive semi-definite matrix \cite{OptBook}. We have already seen that the first derivative is simply $\vx-f(\vx)$, which leads to the conclusion that the second derivative is given by $\bI - \nabla_{\vx} f(\vx)$.

As already mentioned earlier, in the context of some algorithms such as the NLM and the K-SVD, this is associated with the Laplacian $\bI-\bW(\vx)$, and it is positive semi-definite if $\bW$ has all its eigenvalues in the range\footnote{We could in fact allow negative eigenvalues for $\bW$, but this is unnatural in the context of denoising.} $[0,1]$. This is indeed the case for the NLM filter \cite{NLM1}, the K-SVD-denoising algorithm \cite{LapReg9}, and many other denoising engines.

In the wider context of general image denoising engines, convexity is assured if the Jacobian $\nabla_{\vx} f(\vx)$ of the denoising algorithm is stable, as indeed required in Condition 2 in Section \ref{sec:Engine}, $\eta(\nabla_{\vx} f(\vx) ) \le 1$. In this case we have that $\rho_L(\cdot)$ is convex, and this implies that if the log-likelihood expression is convex as well, the proposed scheme is guaranteed to converge to the global optimum of our cost function in Equation (\ref{eqn:BayesianMAP0}). In this respect the proposed algorithm is superior to the $P^3$ scheme in its most general form, which at best is known to get to a stable-point \cite{P3,Follows11}. Furthermore, this result may seem similar to the one posed in  \cite{Follows2,PPP-Convergence}, as our two conditions for global convergence are homogeneity and passivity of $f$, while in these papers the requirements are passivity of $f$, along with a convex energy functional whom $f$ minimizes. The second condition -- having a convex origin to derive $f(\vx)$ -- is in fact more restrictive than demanding homogeneity, as it is unclear which of the known denoisers meet this requirement.

%%%%%%%%%%%%%%%%%%%%%%%%%%%%%%%%%%%%%%%%%%%%%%%%%%%%%%%%%%%%%%%%%%%%%%%%%%%%%%%%

\subsection{An Alternative Prior}

In Section \ref{sec:Algorithm} we motivated the choice of the proposed prior by the desire to characterize the unknown image $\vx$ as one that is not affected by the denoising algorithm, namely, $\vx \approx f(\vx)$. Rather than taking the route proposed, we could have suggested a prior of the form
\begin{eqnarray}
\rho_Q(\vx) = \| \vx - f(\vx) \|_2^2.
\end{eqnarray}
This prior term makes sense intuitively, being based on the same desire to see the denoising residual being small. Indeed, this choice is somewhat related to the option we chose since
\begin{eqnarray}
\rho_Q(\vx) = \| \vx - f(\vx) \|_2^2 = \rho_L(\vx) + f(\vx)^T ( f(\vx)- \vx),
\end{eqnarray}
suggesting a symmetrization of our own expression.

In order to understand the deeper meaning of this alternative, we resort again to the pseudo-linear denoisers, for which this prior is nothing but
\begin{eqnarray}
\rho_Q(\vx) = \| \vx - \bW(\vx) \vx \|_2^2 = \vx^T (\bI - \bW(\vx))^T(\bI - \bW(\vx)) \vx.
\end{eqnarray}
This means that rather than regularizing with the Laplacian, we do so with its square. While this is a worthy possibility which has been considered in the literature under the term ``fourth order regularization'' \cite{FourthOrder}, it is known to be more delicate. We leave this and other possibilities of formulating the regularization with the use of $f(\vx)$ for future work.

%%%%%%%%%%%%%%%%%%%%%%%%%%%%%%%%%%%%%%%%%%%%%%%%%%%%%%%%%%%%%%%%%%%%%%%%%%%%%%%%

\subsection{When is Plug-and-Play-Prior = RED ?}

In Section \ref{sec:Algorithm} we described the use of ADMM as one of the possible avenues for handling our proposed regularization. When handling the inverse problem posed in Equation (\ref{eqn:BayesianMAPLap2}) with ADMM, we have shown that the only difference between this and the $P^3$ scheme resides in the update stage for $\vv$. Here we aim to answer the following question: Assuming that the numerical algorithm used is indeed the ADMM, under what conditions would the two methods ($P^3$ and ours) become equivalent? The answer to this question resides in the optimization task for updating $\vv$, which is a denoising task. Thus, purifying this question, our goal is to find conditions on $f(\cdot)$ and $\lambda$ such that the two treatments of this update stage coincide. Starting from our approach, we would seek the solution of
\begin{eqnarray}
{\hat \vx} = \argmin{\vx}  \frac{\beta}{2}\|\vx - \vy\|_2^2 + \frac{\lambda}{2} \vx^T\left(\vx-f(\vx)\right),
\end{eqnarray}
or, putting it in terms of nulling the gradient of this energy, require
\begin{eqnarray}
 \beta (\vx - \vy) + \lambda (\vx-f(\vx)) =\vzero.
\end{eqnarray}
The ${\widehat \vx}$ that is the solution of this equation is our updated image. On the other hand, the $P^3$ scheme would propose to simply compute\footnote{A delicate matter not considered here is that $P^3$ may apply $\frac{1}{c}f(c\vy)$ in order to tune to a specific noise level. We assume $c=1$ for simplicity.} ${\widehat \vx}=f(\vy)$ as a replacement to this minimization task. Therefore, for the two methods to coincide, we should demand that the gradient expression posed above is solved for the choice of the $P^3$ scheme, namely,
\begin{eqnarray}
 \beta (f(\vy) - \vy)  + \lambda (f(\vy)-f(f(\vy))) =\vzero,
\end{eqnarray}
or posed slightly different,
\begin{eqnarray}
f(\vy)-f(f(\vy)) = \frac{\beta}{\lambda} ( \vy - f(\vy)).
\end{eqnarray}
This means that the denoising residual should remain the same (up to a constant) for the first activation of the denoising engine $\vy - f(\vy)$, and the second one applied on the filtered image $f(\vy)$.

In order to get a better intuition towards this result, let's return to the pseudo-linear case, $f(\vy) = \mW \vy$ with the assumption that $\mW$ is a fixed and diagonalizable matrix. Plugged into the above condition, this gives
\begin{eqnarray}
\mW \vy -\mW^2\vy = \frac{\beta}{\lambda} ( \vy - \mW\vy ),
\end{eqnarray}
or posed differently,
\begin{eqnarray}
\left(\frac{\beta}{\lambda}\mI - \mW\right) \left(\mI -\mW\right)\vy =\vzero.
\end{eqnarray}
As the above equation should hold true for any image $\vy$, we require
\begin{eqnarray}
\left(\frac{\beta}{\lambda}\mI - \mW\right) \left(\mI -\mW\right)=\vzero.
\end{eqnarray}
Without loss of generality, we can assume that $\mW$ is diagonal, after multiplying the above equation from the left and right by the diagonalizing matrix.  With this simplification in mind, we now consider the eigenvalues of $\mW$, and the above equation implies that exact equivalence between our scheme and the $P^3$ one is obtained only if our denoising engine has eigenvalues that are purely $1$'s, or $\beta/\lambda$. Clearly, this is a very limiting case, which suggests that for all other cases, the two methods are likely to differ.

Interestingly, the above analysis is somewhat related to the one given in \cite{Follows1}. Both \cite{Follows1} and our treatment assume that the actual applied denoising engine is $f(\vy) =\mW \vy$ within the ADMM scheme. While we ask for the conditions on $\mW$ to fit our regularization term $\vx^T (\vx - \mW \vx)$, the author of \cite{Follows1} seeks the actual form of the prior to match this step, reaching the conclusion that the prior should be $\vx^T (\mI - \mW) \mW^{\dagger} \vx$. Bearing in mind that the conditions we get for the equivalence between the two methods are too restricting and rarely met, the result in \cite{Follows1} shows the actual gap between the two methods: While we regularize with the expression $\vx^T (\mI - \mW) \vx$, an equivalence takes place only if the $P^3$ modifies this to involve $\mW^{\dagger}$, getting a far more complicated and less natural term.

Just before we conclude this section, we turn briefly to discuss the computational complexity of the proposed algorithm and  its relation to the complexity of the $P^3$ scheme. Put very simply, RED and $P^3$ are roughly of the same computational cost. This is the case when RED is deployed via ADMM and assuming only one iteration in the update of $\vv$, as shown above. Similarly, when using the fixed-point option, RED has the same cost as $P^3$ per iteration.

To conclude, we must state  that this paper is about a more general framework rather than a comparison to the $P^3$. Indeed, one could consider this work as an attempt to provide more solid mathematical foundations for methods like the $P^3$. In addition, when comparing $P^3$ and RED, one can identify several major differences that are far more central than the complexity issue, such as (1) a lack of a clear objective function that $P^3$ serves, while our scheme has a very well-defined penalty; (2) the inability to claim much in terms of convergence of the $P^3$, while our penalty is shown to be convex; (3) the complications of tuning the $P^3$ algorithm, which is very different from the experience we show with RED.

%%%%%%%%%%%%%%%%%%%%%%%%%%%%%%%%%%%%%%%%%%%%%%%%%%%%%%%%%%%%%%%%%%%%%%%%%%%%%%%%
%%%%%%%%%%%%%%%%%%%%%%%%%%%%%%%%%%%%%%%%%%%%%%%%%%%%%%%%%%%%%%%%%%%%%%%%%%%%%%%%

\section{Results}\label{sec:Results}

In this section we compare the performance of the proposed framework to the $ P^3 $ approach, along with various other leading algorithms that are designed to tackle the image deblurring and super-resolution problems. To this end, we plug two substantially different denoising algorithms into the proposed scheme. The first is the (simple) \emph{median filter}, which surprisingly turns out to act as a reasonable regularizer to our ill-posed inverse problems. This option is brought as a core demonstration of the idea that an arbitrary denoiser can be deployed in RED without difficulties. The second denoising engine we use is the state-of-the-art Trainable Nonlinear Reaction Diffusion (TNRD) \cite{chen2015trainable} method. This algorithm trains a nonlinear reaction-diffusion model in a \emph{supervised} manner. As such, in order to treat different restoration problems, one should re-train the underlying model for every specific task -- something we aim to avoid. In the experiments below we build upon the published pre-trained model by the authors of TNRD, tailored to denoise images that are contaminated by white Gaussian noise with a fixed\footnote{In order to handle an arbitrary noise-level, $ \sigma_f $, we rely on the following relation $ f_{\sigma_f}(\vy)~=~\frac{1}{c}f_5(c\cdot\vy) $, where $ c=5/\sigma_f $.} noise-level, which is equal to $5$. Leveraging this, we show how state-of-the-art deblurring and super-resolution results can be achieved simply by integrating the TNRD denoiser in RED. In all the experiments that follow, the parameters were {manually} set in order to enable each method to get its best possible results over the subset of images tested.

%%%%%%%%%%%%%%%%%%%%%%%%%%%%%%%%%%%%%%%%%%%%%%%%%%%%%%%%%%%%%%%%%%%%%%%%%%%%%%%%

\subsection{Image Deblurring}

In order to have a fair comparison to previous work, we follow the synthetic non-blind deblurring experiments conducted in the state-of-the-art work that introduced the Non-locally Centralized Sparse Representation (NCSR) algorithm \cite{NCSR}, which combines the self-similarity assumption \cite{NLM1} with the sparsity-inspired model \cite{Elad_Book}. More specifically, we degrade the test images, supplied by the authors of NCSR, by convolving them with two commonly used point spread functions (PSF); the first is a $ 9 \times 9 $ uniform blur, and the second is a 2D Gaussian function with a standard deviation of $ 1.6 $. In both cases, an additive Gaussian noise with $ \sigma = \sqrt{2} $ is then added to the blurred images. Similarly to NCSR, restoring an RGB image is done by converting it to the YCbCr color-space, applying the deblurring algorithm on the luminance channel only, and then converting the result back to the RGB domain.

Table \ref{tab:deb_results} provides the restoration performance of the three RED schemes -- the steepest-descent (SD), the ADMM, and the fixed-point (FP) methods -- along with the results of the\footnote{We note that $P^3$ using TNRD has never appeared in an earlier publication, and it is brought here in order to let $P^3$ perform as best as it possibly can.} $ P^3 $, the state-of-the-art NCSR and IDD-BM3D \cite{DenGen4}, and two additional  baseline deblurring methods \cite{beck2009fast,dong2011image}. For brevity, only the steepest-descent scheme is presented when considering the basic median filter as a denoiser. The performance is evaluated using the Peak Signal to Noise Ratio (PSNR) measure, higher is better, computed on the luminance channel of the ground-truth and the estimated image. The parameters of the proposed approach, as well as the ones of the $ P^3 $, are tuned to achieve the best performance on this dataset; in the case of the TNRD denoiser, these are depicted in Table \ref{tab:params_laplacian_deblurring} and \ref{tab:params_ppp_deblurring}, respectively. In the setting of the median filter, which extracts the median value of a $ 3\times3 $ window, we choose to run the suggested steepest-descent scheme for $ N=400 $ iterations with $ \lambda = 0.12 $ for the uniform PSF, and $ N=200 $ with $ \lambda = 0.225 $ for the Gaussian PSF.

\begin{table}[htbp]
	\centering
	\tiny
	\setlength{\tabcolsep}{4.5pt}																	
	\renewcommand{\arraystretch}{0.7}
	\renewcommand{\tabcolsep}{6.1pt}
	\begin{tabular}{|c||c|c|c|c|c|c|c|c|c|c|c|}
		\hline
		\textbf{Image} & \textsf{Butterfly} & \textsf{Boats} & \textsf{C. Man} & \textsf{House} & \textsf{Parrot} & \textsf{Lena}  & \textsf{Barbara} & \textsf{Starfish} & \textsf{Peppers} & \textsf{Leaves} & \textbf{Average} \\
		\hline
		\multicolumn{12}{|c|}{\textbf{Deblurring: Uniform kernel, $ \sigma=\sqrt{2} $}} \\
		\hline
		Total Variation \cite{beck2009fast} & 28.37 & 29.04 & 26.82 & 31.99 & 29.11 & 28.33 & 25.75 & 27.75 & 28.43 & 26.49 & 28.21 \\
		IDD-BM3D \cite{DenGen4} & 29.21 & \textbf{31.20} & 28.56 & \textbf{34.44} & 31.06 & 29.70 & \textbf{27.98} & 29.48 & 29.62 & 29.38 & 30.06 \\
		ASDS-Reg \cite{dong2011image} & 28.70  & 30.80  & 28.08 & 34.03 & 31.22 & 29.92 & 27.86 & 29.72 & 29.48 & 28.59 & 29.84 \\
		NCSR  & 29.68 & 31.08 & 28.62 & 34.31 & \textbf{31.95} & 29.96 & 28.10 & 30.28 & 29.66 & 29.98 & 30.36 \\
		$ P^3 $-TNRD & 30.32 & \textbf{31.19} & 28.73 & 33.90 & 31.86 & \textbf{30.13} & 27.21 & 30.27 & \textbf{30.11} & 30.08 & 30.38 \\
		RED: SD-Median Filter & 26.10 & 28.03 & 25.57 & 29.81 & 28.67 & 27.29 & 25.62 & 27.84 & 27.40 & 25.45 & 27.18 \\
		RED: SD-TNRD & 30.20 & \textbf{31.20} & 28.67 & 33.83 & 31.62 & 29.98 & 27.35 & 30.47 & 30.10 & 29.72 & 30.31 \\
		RED: ADMM-TNRD & \textbf{30.40}	& 31.12	& 28.71	& 33.77	& 31.86	& 30.03	& 27.27	& \textbf{30.58}	& \textbf{30.11}	& \textbf{30.12}	& \textbf{30.40} \\
		RED: FP-TNRD & \textbf{30.41} & 31.12 & \textbf{28.76} & 33.76 & 31.83 & 30.02 & 27.27 & \textbf{30.57} & \textbf{30.12} & \textbf{30.13} & \textbf{30.40} \\
		\hline
		\multicolumn{12}{|c|}{\textbf{Deblurring: Gaussian kernel, $ \sigma=\sqrt{2} $}} \\
		\hline
		Total Variation \cite{beck2009fast} & 30.36 & 29.36 & 26.81 & 31.50 & 31.23 & 29.47 & 25.03 & 29.65 & 29.42 & 29.36 & 29.22 \\
		IDD-BM3D \cite{DenGen4} & 30.73 & \textbf{31.68} & 28.17 & \textbf{34.08} & 32.89 & 31.45 & 27.19 & 31.66 & 29.99 & 31.4  & 30.92 \\
		ASDS-Reg \cite{dong2011image} & 29.83 & 30.27 & 27.29 & 31.87 & 32.93 & 30.36 & 27.05 & 31.91 & 28.95 & 30.62 & 30.11 \\
		NCSR \cite{NCSR} & 30.84 & 31.49 & 28.34 & 33.63 & 33.39 & 31.26 & \textbf{27.91} & 32.27 & 30.16 & 31.57 & 31.09 \\
		$ P^3 $-TNRD & \textbf{31.73} & \textbf{31.67} & 28.08 & {33.95} & \textbf{33.43} & \textbf{31.52} & 27.11 & \textbf{32.71} & \textbf{30.94} & \textbf{32.18} & \textbf{31.33} \\
		RED: SD-Median Filter & 29.02 & 30.01 & 26.45 & 31.59 & 31.32 & 30.00 & 25.02 & 30.29 & 28.53 & 28.69 & 29.09 \\
		RED: SD-TNRD & 31.57 & 31.53 & 28.31 & 33.71 & 33.19 & 31.47 & 26.62 & 32.46 & 29.98 & 31.95 & 31.08 \\
		RED: ADMM-TNRD &  31.66	& 31.55	& 28.31	& 33.73	& 33.33	& 31.40	& 26.76	& 32.49	& 30.48	& 31.93	& 31.16 \\
		RED: FP-TNRD & 31.66 & 31.55 & \textbf{28.38} & 33.74 & 33.33 & 31.39 & 26.76 & 32.49 & 30.51 & 31.93 & 31.17 \\
		\hline
	\end{tabular}
	\caption{Deblurring results measured in PSNR [dB] and evaluated on the set of images provided by the authors of NCSR \cite{NCSR}. The $ P^3 $ and RED build upon the TNRD \cite{chen2015trainable} as the denoising engine. We also provide the results obtained by integrating the median filter with the steepest-descent RED scheme. {PSNR scores being less than $ 0.01 $[dB] away from the highest result are highlighted. Note that the $ P^3 $ does not converge when setting the TNRD to be the denoising algorithm. Therefore, we run the $ P^3 $ for a fixed number of iterations, chosen to achieve the best PSNR on average (otherwise the restoration quality would be significantly inferior). Please refer to Figure \ref{fig:red_vs_ppp} and Section \ref{sec:Parameters} for more details regarding the sensitivity of the $ P^3 $ to the choice of parameters. }}
	\label{tab:deb_results}
\end{table}

\begin{table}[htbp]
	\centering		
	\scriptsize
	\renewcommand{\arraystretch}{0.8}
	\renewcommand{\tabcolsep}{9pt}
	\begin{tabular}{|c|c|c|c|c|}
		\hline
		\multirow{3}{*}{\textbf{PSF}} & \multicolumn{4}{c|}{\textbf{Proposed approach: Deblurring}} \\
		\cline{2-5}
		& \multirow{2}{*}{\textbf{Parameter}} & \textbf{Steepest} & \textbf{Fixed} & \multirow{2}{*}{\textbf{ADMM}} \\
		&                            & \textbf{Descent}   & \textbf{Point} &                       \\
		\hline \hline
		\multirow{6}{*}{\rotatebox[origin=c]{90}{Uniform}} & N     & 1500  & 200   & 200 \\
		& $ \sigma_f $ & $3.25 $ & $3.25 $ & $3.25 $ \\
		& $ \lambda $ & 0.02  & 0.02  & 0.02 \\
		& $ m_1 $    & --     & closed-form using FFT   & closed-form using FFT \\
		& $ m_2 $    & --     & --     & 1 \\
		& $ \beta $  & --     & --     & 0.001 \\
		\hline \hline
		\multirow{6}{*}{\rotatebox[origin=c]{90}{Gaussian}} & N     & 1500  & 200   & 200 \\
		& $ \sigma_f $ & $4.1 $ & $4.1 $ & $4.1 $ \\
		& $ \lambda $ & 0.01  & 0.01  & 0.01 \\
		& $ m_1 $    & --     & closed-form using FFT   & closed-form using FFT \\
		& $ m_2 $    & --     & --     & 1 \\
		& $ \beta $  & --     & --     & 0.001 \\
		\hline
	\end{tabular}%
	\caption{The set of parameters being used in our framework, leading to the deblurring results reported in Table~\ref{tab:deb_results} when plugging the TNRD \cite{chen2015trainable} denoiser.}
	\label{tab:params_laplacian_deblurring}%
\end{table}%

\begin{table}[htbp]
	\centering
	\scriptsize
	\renewcommand{\arraystretch}{0.8}
	\renewcommand{\tabcolsep}{9pt}
	\begin{tabular}{|c|c|c|}
		\hline
		\multirow{2}{*}{\textbf{PSF}} & \multicolumn{2}{c|}{$ P^3 $\textbf{: Deblurring}} \\
		\cline{2-3}
		& \textbf{Parameter} & \textbf{Value} \\
		\hline \hline
		\multirow{6}{*}{{\rotatebox[origin=c]{90}{Uniform}}} & N     & 200 \\
		& $ \alpha $ & 1.02 \\
		& $ \beta_0 $ & 0.0007 \\
		& $ \beta_k $ & $ \alpha^k \cdot \beta_0 $ \\
		& $ \lambda $ & $ 512 \cdot \beta_0 $ \\
		& $ \sigma_f $ & $ \sqrt{\lambda/\beta_k} $ \\
		\hline \hline
		\multirow{6}{*}{\rotatebox[origin=c]{90}{Gaussian}} & N     & 200 \\
		& $ \alpha $ & 1.02 \\
		& $ \beta_0 $ & 0.0007 \\
		& $ \beta_k $ & $ \alpha^k \cdot \beta_0 $ \\
		& $ \lambda $ & $ 320 \cdot \beta_0 $ \\
		& $ \sigma_f $ & $ \sqrt{\lambda/\beta_k} $ \\
		\hline
	\end{tabular}%
	\caption{{The set of parameters being used by the $ P^3 $ method, leading to the deblurring results reported in Table~\ref{tab:deb_results} when plugging the TNRD \cite{chen2015trainable} denoiser.}}
	
	\label{tab:params_ppp_deblurring}%
\end{table}%

\begin{figure}
	\begin{center}
		\subfigure[RED, Denoising Engine: Median Filter]{\includegraphics[width=0.48\linewidth]{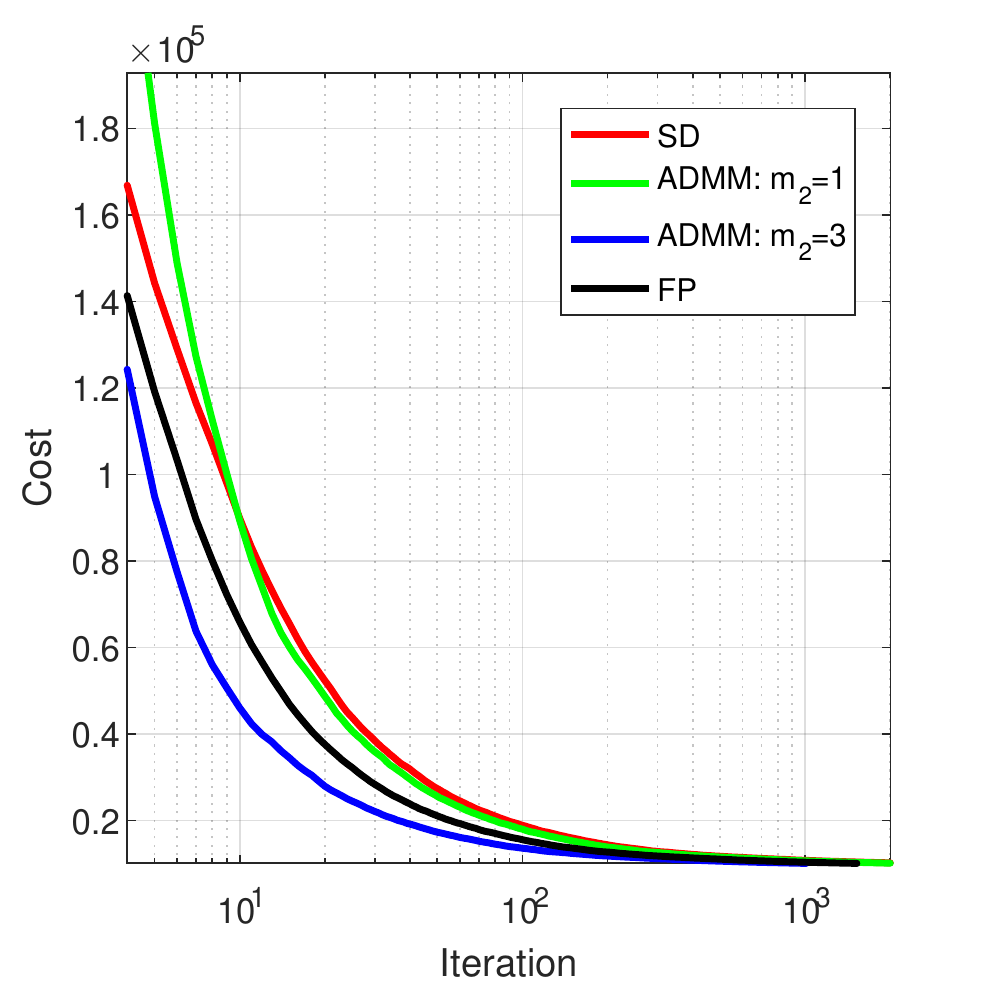}}
		\subfigure[RED, Denoising Engine: TNRD]{\includegraphics[width=0.48\linewidth]{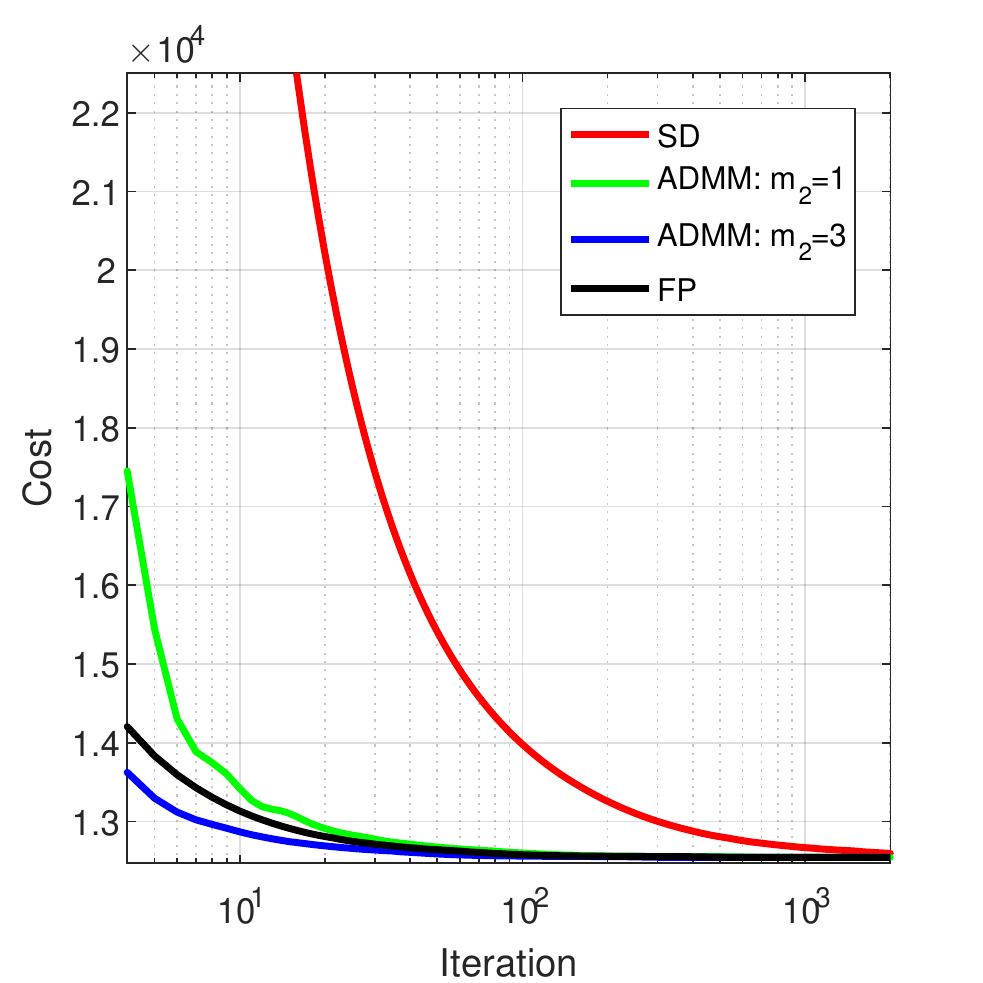}}
	\end{center}
	\caption{{An illustration of the convergence of RED using three proposed numerical schemes -- the steepest-descent (red), the ADMM with $m_2 = 1$ (green), the ADMM with $m_2 = 3$ (blue) and the fixed-point (black). These are applied on the image \textsf{Leaves}, degraded by a Gaussian PSF, when two denoising engines are tested: (a) the median filter, and (b) TNRD \cite{chen2015trainable}.}}
	\label{fig:convergence_deb_gauss}
\end{figure}

\begin{figure*}
	\begin{center}
		\subfigure[Ground Truth]{\includegraphics[]{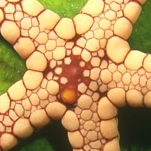}}
		\subfigure[Input 20.83dB]{\includegraphics[]{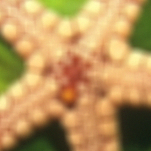}}
		\subfigure[RED: SD-Median filter 25.87dB]{\includegraphics[]{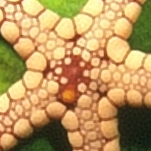}}
		\subfigure[NCSR 28.39dB]{\includegraphics[]{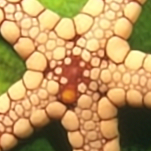}}
		\subfigure[$ P^3 $-TNRD 28.43dB]{\includegraphics[]{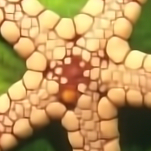}}
		\subfigure[RED: FP-TNRD 28.82dB]{\includegraphics[]{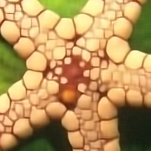}}
	\end{center}
	\caption{Visual comparison of deblurring a cropped area from the image \textsf{Starfish}, degraded by a uniform PSF, along with the corresponding PSNR [dB] score.}
	\label{fig:comp_vis_deb_ones}
\end{figure*}

\begin{figure*}
	\begin{center}
		\subfigure[Ground Truth]{\includegraphics[]{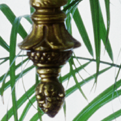}}
		\subfigure[Input 21.40dB]{\includegraphics[]{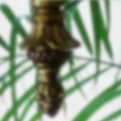}}
		\subfigure[RED:SD-Median filter 27.87dB]{\includegraphics[]{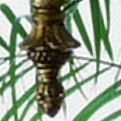}}
		\subfigure[NCSR 30.03dB]{\includegraphics[]{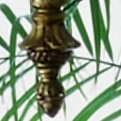}}
		\subfigure[$ P^3 $-TNRD 30.36dB]{\includegraphics[]{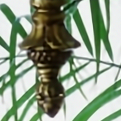}}
		\subfigure[RED: ADMM-TNRD 30.40dB]{\includegraphics[]{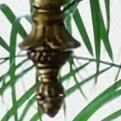}}
	\end{center}
	\caption{{Visual comparison of deblurring a cropped area from the image \textsf{Leaves}, degraded by a Gaussian PSF, along with the corresponding PSNR [dB] score.}}
	\label{fig:comp_vis_deb_gauss}
\end{figure*}

Several remarks are to be made with regard to the obtained results. When the image is degraded by a Gaussian blur kernel, integrating the median filter in the proposed framework leads to a surprising restoration performance that is similar to the total variation deblurring \cite{beck2009fast}. Furthermore, by choosing the state-of-the-art TNRD to be our denoising engine we achieve results that are competitive with the impressive NCSR and IDD-BM3D methods, which are specifically designed to tackle the deblurring task. Notice that the three versions of the proposed framework obtain a similar PSNR score. However, while the ADMM and the fixed-point variants are of similar complexity, the steepest-descent requires many more steps to converge and thereby more applications of the denoiser. As our last observation, based on the obtained results we conclude that the proposed approach is equivalent in quality to the alternative $ P^3 $ framework. {However, tuning the parameters of the proposed algorithm is significantly simpler than the ones of the $ P^3 $; while in the $ P^3 $ the parameters should be modified throughout the iterations, in our approach these are always fixed (refer to Section \ref{sec:Parameters} for a broader discussion).}

{An illustration of the convergence of the proposed approach using the three different numerical schemes (steepest-descent, ADMM, and fixed-point) and the two denoisers (median filter, and TNRD) is given in Figure~\ref{fig:convergence_deb_gauss}. As can be observed, the three algorithms indeed converge, but at different rates; the steepest-descent is the slowest, while the ADMM is the fastest. The fixed-point strategy is slightly slower than the ADMM alternative, but requires only one application of the denoiser per iteration while the ADMM demands $ m_2 = 3$ such operations. However, when comparing the fixed-point to the ADMM with the setting $ m_2 = 1 $ (now both have the same computational cost) we observe that the fixed-point is faster.}

The above discussion is also supported visually in Figure \ref{fig:comp_vis_deb_ones} and \ref{fig:comp_vis_deb_gauss}, comparing the proposed method to the $ P^3 $ and the NCSR both for uniform and Gaussian PSF. As can be seen, by plugging the median filter into RED we improve the quality of the blurry image, yet the gap in performance between this simplistic approach and the state-of-the-art is substantial. Once relying on the TNRD, we obtain an efficient deblurring machine that has comparable results to $ P^3 $, and both are competitive or even slightly better than the NCSR.

%%%%%%%%%%%%%%%%%%%%%%%%%%%%%%%%%%%%%%%%%%%%%%%%%%%%%%%%%%%%%%%%%%%%%%%%%%%%%%%%

\subsection{Image Super-Resolution}

Similarly to the previous subsection, we imitate the super-resolution experiments done in \cite{NCSR}. To this end, a low-resolution image is generated by blurring the ground-truth high-resolution one with a $ 7\times 7 $ Gaussian blur kernel with standard deviation $1.6$, followed by down-sampling by a factor of $ 3 $ in each axis. Next, white Gaussian noise of standard deviation $5$ is added to the low-resolution images. The upscaling of an RGB image is done by transforming it first to the YCbCr color-space, super-resolving the luminance channel using the proposed approach (or the baseline methods), while the chroma channels are upscaled by bicubic interpolation. Lastly, the outcome is converted back to the RGB color-space.

In terms of PSNR, Table~\ref{tab:sr_results} presents the restoration performance of the three variants of the proposed approach in addition to the ones of the $ P^3 $, the NCSR and the ASDS-Reg \cite{dong2011image} algorithms. Similarly to the deblurring case, the PSNR is computed on the luminance channel only. In the case of the TNRD denoiser, the parameters that are used in our approach and in the $ P^3 $ are listed in Table~\ref{tab:params_laplacian_sr} and \ref{tab:params_ppp_sr}, respectively. In the simpler case of the median filter (defined by a $ 3\times3 $ window), the number of iterations of the proposed steepest-descent algorithm is set to $ N=50 $ with a parameter $ \lambda = 0.0325 $.

\begin{table}[htbp]
	\centering
	\tiny
	\renewcommand{\arraystretch}{0.8}
	\renewcommand{\tabcolsep}{7.5pt}
	\begin{tabular}{|c||c|c|c|c|c|c|c|c|c|c|}
		\hline
		\multicolumn{11}{|c|}{\textbf{Super-Resolution, scaling = 3, $ \sigma=\sqrt{2} $}} \\
		\hline
		\textbf{Image} & \textsf{Butterfly} & \textsf{flower} & \textsf{Girl}  & \textsf{Parth.} & \textsf{Parrot} & \textsf{Raccoon} & \textsf{Bike}  & \textsf{Hat}   & \textsf{Plants} & \textbf{Average} \\
		\hline
		Bicubic & 20.74 & 24.74 & 29.61 & 24.04 & 25.43 & 26.20 & 20.75 & 27.00 & 27.62 & 25.13 \\
		ASDS-Reg \cite{dong2011image} & 26.06 & 27.83 & 31.87 & 26.22 & 29.01 & 28.01 & 23.62 & 29.61 & 31.18 & 28.16 \\
		NCSR  \cite{NCSR} & 26.86 & 28.08 & 32.03 & 26.38 & 29.51 & 28.03 & 23.80 & 29.94 & 31.73 & 28.48 \\
		$P^3$-TNRD & 27.13 & \textbf{28.23} & \textbf{32.08} & 26.50 & \textbf{29.65} & 27.95 & \textbf{24.04} & 30.30 & \textbf{31.78} & 28.63 \\
		RED: SD-Median Filter & 24.44 & 27.24 & 31.13 & 25.80 & 27.76 & 27.65 & 22.89 & 28.69 & 30.24 & 27.32 \\
		RED: SD-TNRD & \textbf{27.37} & \textbf{28.23} & \textbf{32.08} & \textbf{26.54} & 29.43 & \textbf{27.98} & \textbf{24.04} & \textbf{30.36} & \textbf{31.79} & \textbf{28.65} \\
		RED: ADMM-TNRD & 27.22	& \textbf{28.24}	& \textbf{32.08}	& 26.51	& 29.41	& \textbf{27.97}	& 23.96	& \textbf{30.35}	& 31.77	& 28.61 \\
		RED: FP-TNRD & 27.26 & \textbf{28.24} & \textbf{32.08} & 26.52 & 29.42 & \textbf{27.97} & 23.97 & \textbf{30.35} & 31.77 & 28.62 \\
		\hline
	\end{tabular}%
	\caption{Super-resolution results, measured in terms of PSNR [dB] and evaluated on the set of images provided by the authors of NCSR \cite{NCSR}. The $ P^3 $ and ours steepest-descent (SD), ADMM, and fixed-point (FP) methods build upon the TNRD \cite{chen2015trainable} as the denoising engine. We also provide the results obtained by integrating the median filter in the steepest-descent scheme. {PSNR scores being less than $ 0.01 $[dB] away from the highest result are highlighted. Similarly to the deblurring case, the $ P^3 $ does not converge here as well. Therefore, we run it for a fixed number of iterations, manually tuned to achieve the highest PSNR on average. This behavior is demonstrated in detail in Figure \ref{fig:red_vs_ppp}, and discussed in Section \ref{sec:Parameters}. }}
	
	\label{tab:sr_results}%
\end{table}%

\begin{table}[htbp]
	\centering
	\scriptsize
	\renewcommand{\arraystretch}{0.8}
	\renewcommand{\tabcolsep}{9pt}
	\begin{tabular}{|c|c|c|c|}
		\hline
		\multicolumn{4}{|c|}{\textbf{Proposed approach: Super-Resolution}} \\
		\hline
		\multirow{2}{*}{\textbf{Parameter}} & \textbf{Steepest} & \textbf{Fixed} & \multirow{2}{*}{\textbf{ADMM}} \\
		& \textbf{Descent}   & \textbf{Point} &                       \\
		\hline \hline
		N     & 1500  & 200   & 200 \\
		$ \sigma_f $ & $ 3 $ & $ 3 $ & $3 $ \\
		$ \lambda $ & 0.008  & 0.008  & 0.008 \\
		$ m_1 $    & --     & 200 (until convergence)   & 200 (until convergence) \\
		$ m_2 $    & --     & --     & 1 \\
		$ \beta $  & --     & --     & 0.001 \\
		\hline
	\end{tabular}%
	\caption{The set of parameters being used in our framework, leading to the super-resolution results reported in Table~\ref{tab:sr_results} when plugging the TNRD \cite{chen2015trainable} denoiser.}
	
	\label{tab:params_laplacian_sr}%
\end{table}%

\begin{table}[htbp]
	\centering
	\scriptsize
	\renewcommand{\arraystretch}{0.8}
	\renewcommand{\tabcolsep}{9pt}
	\begin{tabular}{|c|c|}
		\hline
		\multicolumn{2}{|c|}{$ P^3 $\textbf{: Super-Resolution}} \\
		\hline
		\textbf{Parameter} & \textbf{Value} \\
		\hline \hline
		N     & 200 \\
		$ \alpha $ & 1.02 \\
		$ \beta_0 $ & 0.001 \\
		$ \beta_k $ & $ \alpha^k \ \beta_0 $ \\
		$ \lambda $ & $ 360 \ \beta_0 $ \\
		$ \sigma_f $ & $ \sqrt{\lambda/\beta_k} $ \\
		\hline
	\end{tabular}%
	\caption{The set of parameters being used in the $ P^3 $ method, leading to the super-resolution results reported in Table~\ref{tab:sr_results}  when plugging the TNRD \cite{chen2015trainable} denoiser.}		
	
	\label{tab:params_ppp_sr}%
\end{table}%

\begin{figure*}
		\begin{center}
		\subfigure[Bicubic 20.68dB]{\includegraphics[]{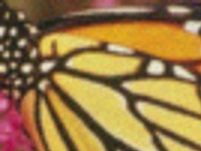}}
		\subfigure[NCSR 26.79dB]{\includegraphics[]{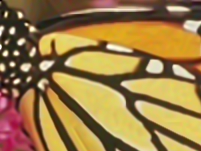}}
		\subfigure[$ P^3 $-TNRD 26.61dB]{\includegraphics[]{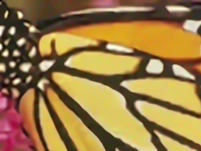}}
		\subfigure[Ours: SD-TNRD 27.39dB]{\includegraphics[]{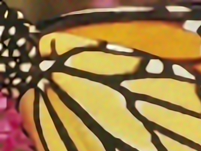}}
	\end{center}
	\caption{{Visual comparison for upscaling by a factor of 3 for the image \textsf{Butterfly}, along with the corresponding PSNR [dB] score.}}
	\label{fig:comp_vis_sr_butterfly}
\end{figure*}

\begin{figure*}
	\begin{center}
		\subfigure[Bicubic 20.44dB]{\includegraphics[]{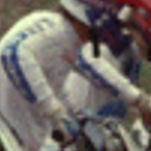}}
		\hfil
		\subfigure[NCSR 22.97dB]{\includegraphics[]{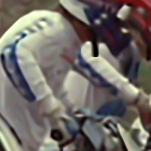}}
		\\
		\subfigure[$ P^3 $-TNRD 23.25dB]{\includegraphics[]{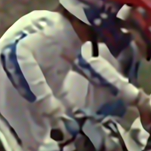}}
		\hfil
		\subfigure[Ours: ADMM-TNRD 23.28dB]{\includegraphics[]{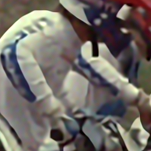}}
	\end{center}
	\caption{{Visual comparison for upscaling by a factor of 3 for the image \textsf{Bike}, along with the corresponding PSNR [dB] score.}}
	\label{fig:comp_vis_sr_bike}
\end{figure*}

Interestingly, when setting the median filter to be our denoising engine we get a 2.19dB improvement (on average) over the bicubic interpolation. Alternatively, when choosing a stronger denoiser such as the TNRD, we achieve state-of-the-art results. Notably, the $ P^3 $ and the three variants of the proposed approach lead to similar restoration performance, consistent with the observation that was made in the context of the deblurring problem. These support once again the claim that our framework is a tangible alternative to the $ P^3 $. Figure \ref{fig:comp_vis_sr_butterfly} and \ref{fig:comp_vis_sr_bike} visually compare the proposed method to the $ P^3 $ and also to the state-of-the-art NCSR. As shown, the three algorithms offer an impressive restoration with sharp and clear edges, complying with the quantitative results which are given in Table \ref{tab:sr_results}.

\subsection{Robustness to the Choice of Parameters} \label{sec:Parameters}

\begin{figure*}
	\begin{center}
		\subfigure[Comparison between RED and $ P^3 $]{\includegraphics[width=0.48\linewidth]{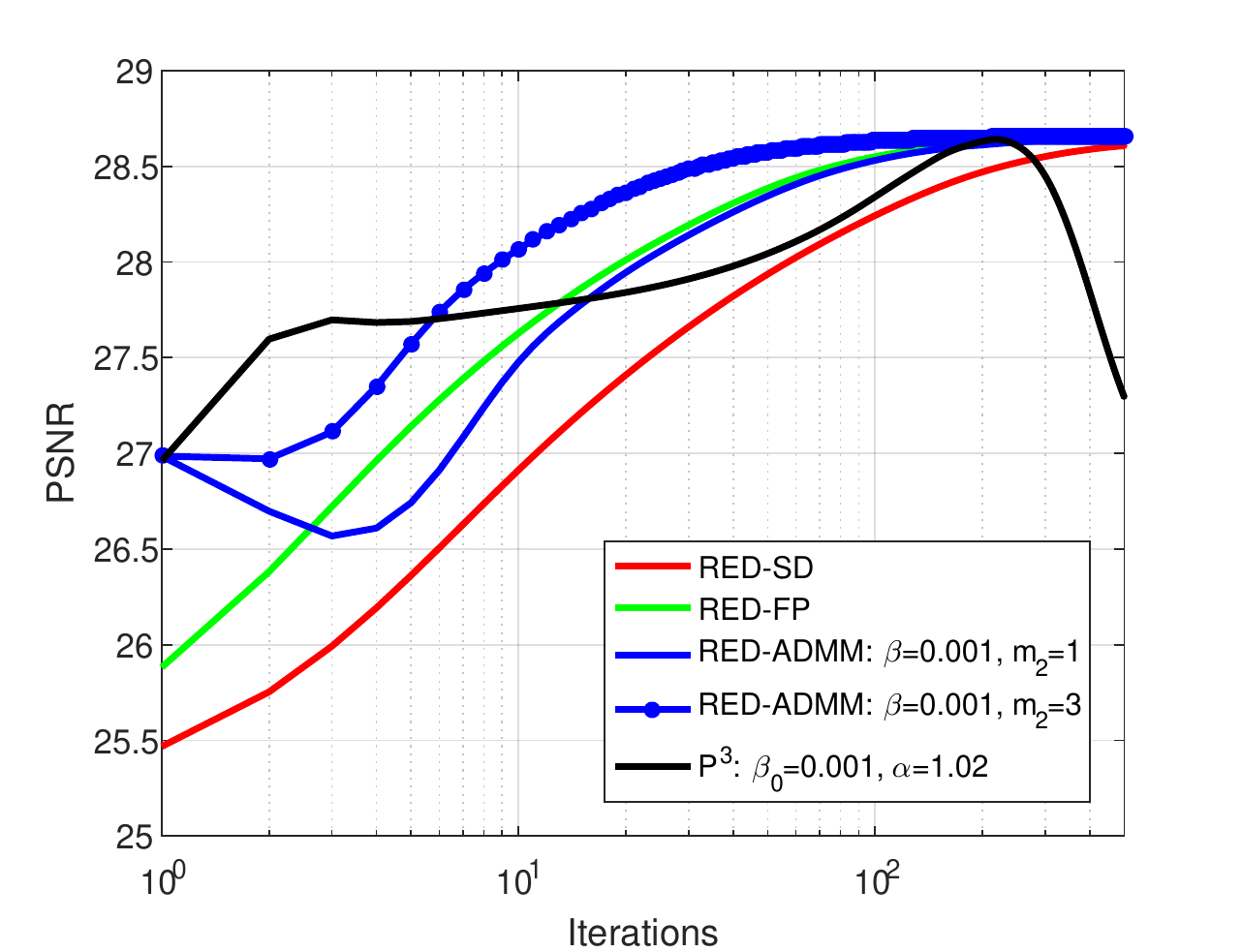}}
		\subfigure[$ P^3 $: Sensitivity to change in $ \beta_0 $ and $ \alpha $]{\includegraphics[width=0.48\linewidth]{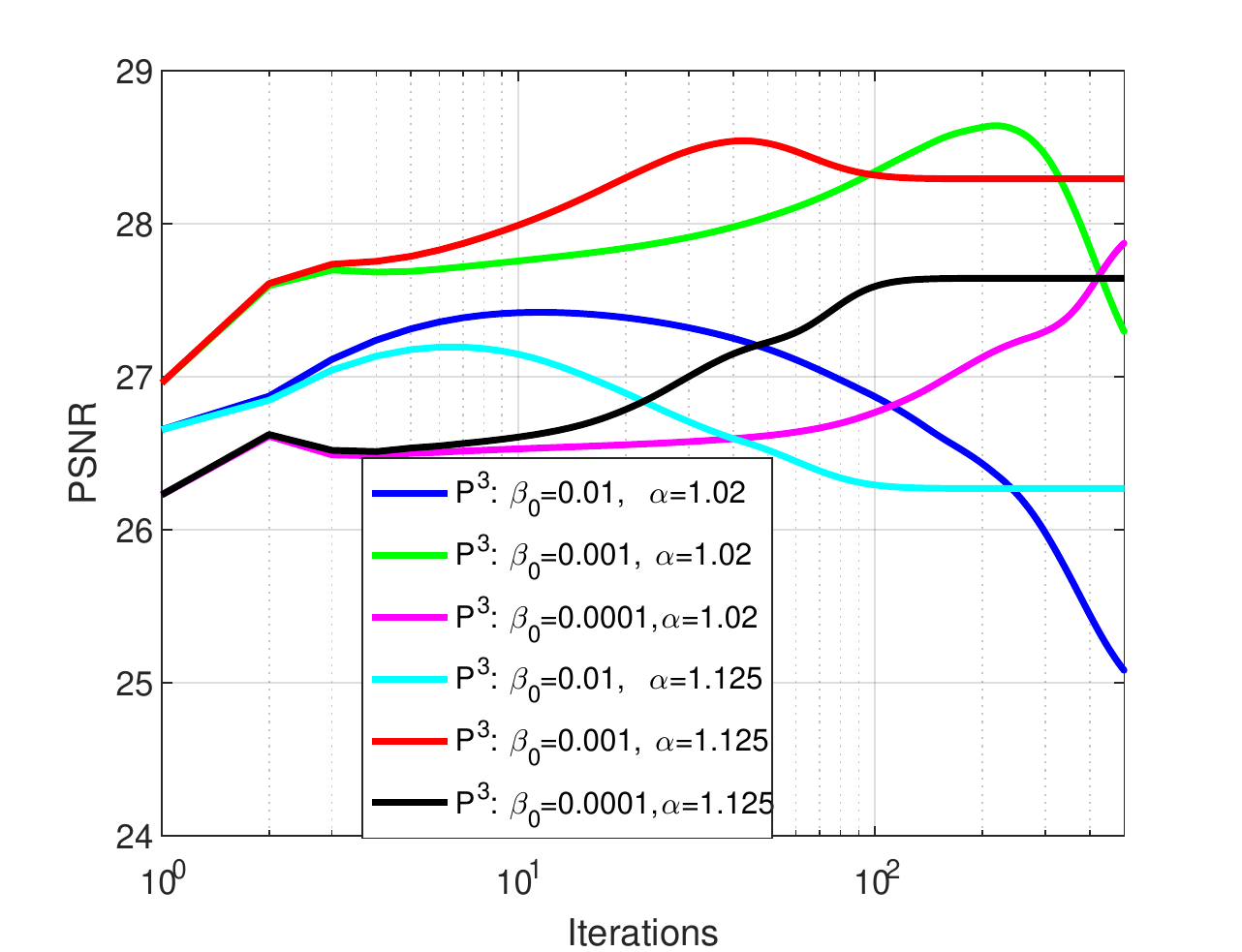}}
		\subfigure[RED: Robustness to change in $ \beta $ and $ m_2 $]{\includegraphics[width=0.48\linewidth]{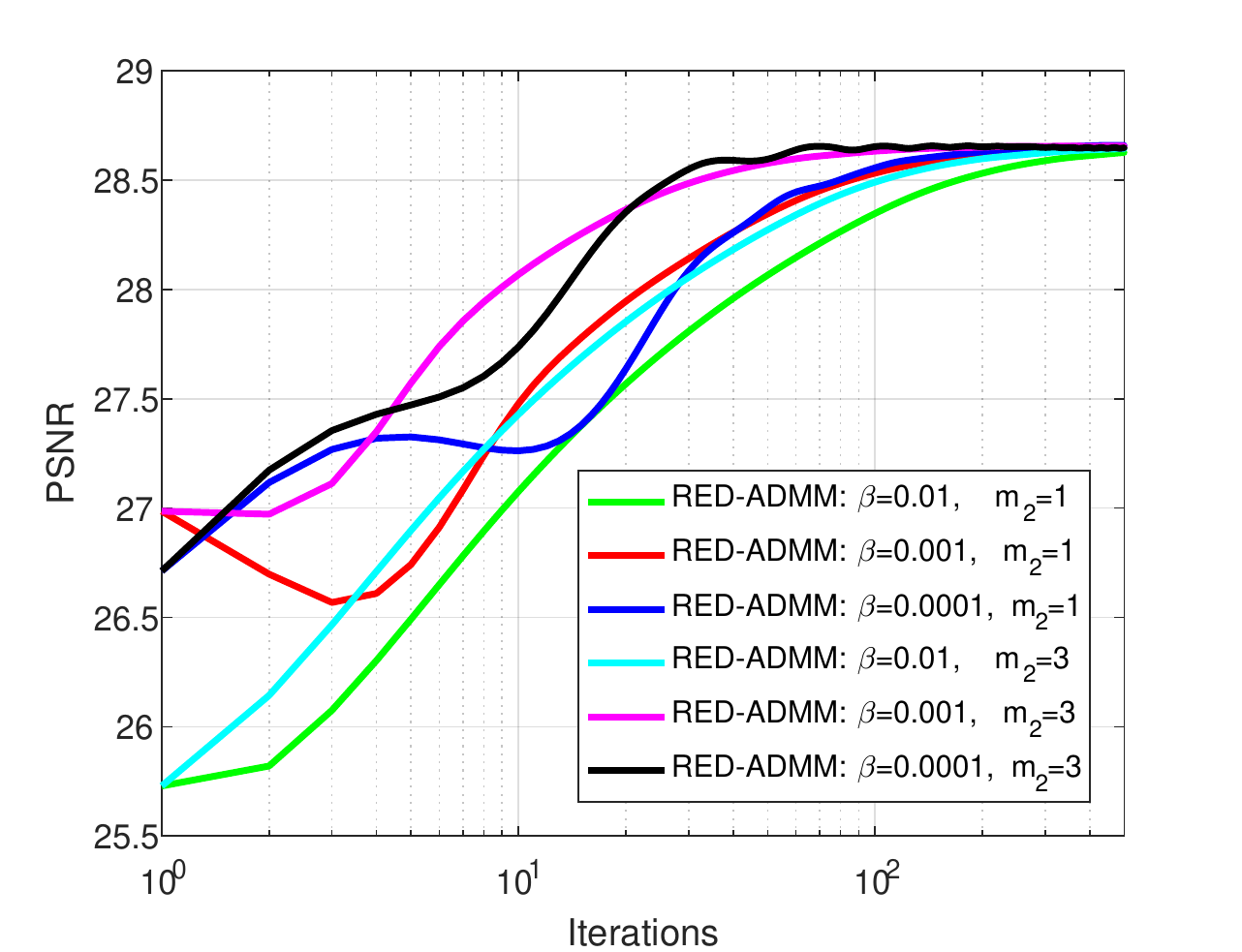}}
		\subfigure[RED: The effect of a change in $ \sigma_f$]{\includegraphics[width=0.48\linewidth]{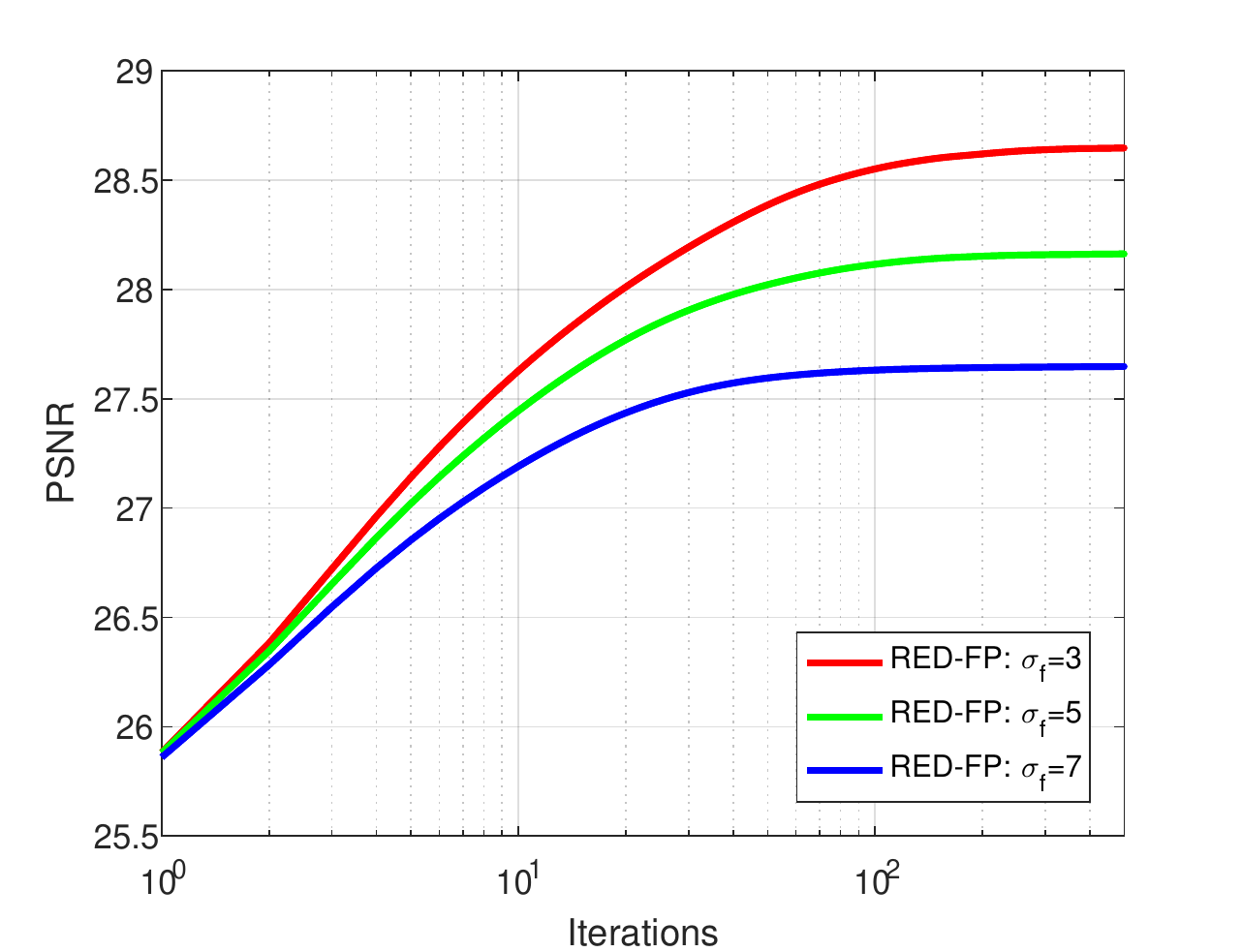}}
	\end{center}
	\caption{{Image super resolution. Average PSNR as a function of the iterations of the different methods. The test images are the ones of Table \ref{tab:sr_results}.}}
	\label{fig:red_vs_ppp}
\end{figure*}

In this subsection we test the robustness of RED to the choice of its parameters, and contrast it to the $ P^3 $. To this end, we choose the single image super-resolution problem as a case-study, described in the previous subsection. Figure \ref{fig:red_vs_ppp} (a) plots the average PSNR obtained by the different approaches as a function of the outer iterations. One can observe that RED (in all its forms) converges to a similar PSNR value. Also, an increase of $ m_2 $ (the number denoising steps within each iteration) leads to an improved rate of convergence of the ADMM. On the other hand, the curve describing the $ P^3 $ shows an unstable behavior and tendency to decrease in the PSNR after the first 200 iterations. Note that no tool has been suggested in the literature so far to automatically stop the $P^3$ for extracting the best performing outcome.

This unstable nature of the $ P^3 $ appears again as we modify the values of $\alpha$ and $\beta_0$. Figure \ref{fig:red_vs_ppp} (b) shows the behavior of $P^3$ for several settings of these two parameters, clearly exhibiting an erratic behavior. One could observe that for specific choices of these two parameters, convergence is obtained, as manifested by the flattened curves. However, this is a fake convergence, caused by a large enough value of $\beta_k$. We stress that, in principle, a change in $ \beta_0 $ is expected to modify the convergence rate of the ADMM, but the steady state outcome should remain the same. However, when observing the curves in Figure \ref{fig:red_vs_ppp} (b), it is clear that this is not the case in the $ P^3 $.

Back to RED, we repeat a similar experiment and test the sensitivity (or better yet the robustness) of the ADMM to the choice of $ \beta $. Figure \ref{fig:red_vs_ppp} (c) shows that different values of $ \beta $ indeed affect the convergence rate (as expected), but the PSNR of the final outcome is always the same. This figure also indicates that the more accurate the solution of Part II in Figure \ref{fig:algoADMM} (obtained by increasing the value of $ m_2 $), the better the convergence rate.

The sensitivity of RED to the choice of $ \sigma_f $ -- the input noise-level to the denoiser -- is depicted in Figure \ref{fig:red_vs_ppp} (d). Notice that the choice of $ \sigma_f $ affects directly the proposed regularizer, given by $ \lambda\rho(\vx,\sigma_f) = \frac{\lambda}{2} \: \vx^T\left[\vx-f_{\sigma_f}(\vx)\right] $. As such, a change in $ \sigma_f $ is expected to modify the objective and thereby the resulting PSNR, as shown empirically in Figure \ref{fig:red_vs_ppp} (d) for the FP method\footnote{One expects that we could tune this parameter using existing techniques such as SURE, but we leave this topic for future research.}. Clearly, a similar behavior is expected to occur when modifying the weight of the regularizer, $ \lambda $, in which we choose to omit from this experimental part for brevity.

%%%%%%%%%%%%%%%%%%%%%%%%%%%%%%%%%%%%%%%%%%%%%%%%%%%%%%%%%%%%%%%%%%%%%%%%%%%%%%%%
%%%%%%%%%%%%%%%%%%%%%%%%%%%%%%%%%%%%%%%%%%%%%%%%%%%%%%%%%%%%%%%%%%%%%%%%%%%%%%%%

\section{Conclusions}\label{sec:Conclusions}

The idea put forward in this paper is strange -- using a denoising engine within the regularization term in order to handle general inverse problems. A surprising outcome of this proposal is the fact that differentiation of the regularization terms remains tractable, still using the very same denoiser engine and {\em not} its derivative. This led us to the proposed scheme, termed Regularization by Denoising (RED). We have shown and discussed various appealing properties of this approach, such as convexity and its relation to advanced Laplacian smoothing. We have contrasted this scheme with the plug-and-play-prior method \cite{P3}, and we have provided experiments that demonstrate the validity of the regularization strategy and the resulting competitive performance.

Is there a wider message in this work? Could it be that one could use more general $f$ functions and still form the proposed regularization? We have seen that all that it takes is the availability of the directional derivative of this function in order to follow all through. Strangely enough, we have shown how even a median filter could fit into this scheme, despite the fact that it does not have any relation to Gaussian denoising. More work is required to investigate the ability to further generalize RED to other and perhaps more daring  regularization functionals.

A key question we have left open at this stage is the setting of the parameter $\sigma_f$. We chose this to be a fixed value, but we did not address the question of its influence on the overall performance, or whether a varying value strategy could lead to a benefit. More work is required here as well.

%%%%%%%%%%%%%%%%%%%%%%%%%%%%%%%%%%%%%%%%%%%%%%%%%%%%%%%%%%%%%%%%%%%%%%%%%%%%%%%%
%%%%%%%%%%%%%%%%%%%%%%%%%%%%%%%%%%%%%%%%%%%%%%%%%%%%%%%%%%%%%%%%%%%%%%%%%%%%%%%%
%%%%%%%%%%%%%%%%%%%%%%%%%%%%%%%%%%%%%%%%%%%%%%%%%%%%%%%%%%%%%%%%%%%%%%%%%%%%%%%%

\appendix
\section*{Appendices}
\section{Can We Mimic any Prior?}
So far we have shown that denoisers $f(\vx)$ can be used to define the powerful family of priors $\vx^T(\vx-f(\vx))$ that can regularize a wide variety of inverse problem. Now, let's consider the reverse question: For a given $\rho(\vx)$, what is the effective $f(\vx)$ behind it (if any)? More specifically, given $\rho(\vx)$, we wish to find $f(\vx)$ such that
\begin{equation} \label{eq:priorform}
\frac{1}{2} \vx^T(\vx-f(\vx)) = \rho(\vx)
\end{equation}
Recall that one of the key conditions we assumed for the denoiser $f(\vx)$ is that it is homogenous of degree $1$,
\begin{equation}
f(c\vx) = c f(\vx).
\end{equation}
This immediately notifies us that the $\rho(\vx)$ we wish to mimic in (\ref{eq:priorform}) must be $2$-homogenous because
\begin{equation}
\rho(c\vx) = \frac{1}{2}(c\vx)^T (c\vx - f(c\vx)) = \frac{1}{2}c^2 \vx^T(\vx - f(\vx)) = c^2 \rho(\vx)
\end{equation}
Of course, not all priors satisfy this condition, but the class of priors that do is wide. For instance, while the total-variation function TV($\vx)=\|\nabla x \|_1$ is only $1$-homogeneous, its square $\|\nabla x \|_1^2$ keeps us in business. More generally, since any norm is {\em by definition} $1$-homogenous, all regularizers of the type $\rho(\vx) = \| \bA x \|_q^{2}$  for $q = 1, 2, \cdots, \infty$ are $2$-homogeneous\footnote{The sparsity-inspired $L_0$ is an exception that can not be treated this way.}:
\begin{equation}
\rho(c\vx) = \| \bA (c\vx) \|_q^{2} = (c\| \bA \vx \|_q)^{2} = c^2\| \bA \vx \|_q^2 = c^2\rho(\vx)
\end{equation}
Squared norms are not the only functions at our disposal. Beyond norms, any $k$-homogenous function (such as homogeneous polynomials of degree $k$) can be made $2$-homogeneous by raising to power $2/k$. As well, order-statistics such as maximum, minimum and median are $1$-homonegous, and can be squared to the same end.

With the above discussion, we move forward with the assumption that we have a $2$-homogeous prior $\rho(\vx)$ in our hands. Let's recall that the task is to find $f(\vx)$ so that
\begin{equation}
\frac{1}{2} \vx^T(\vx-f(\vx)) = \rho(\vx)
\end{equation}
We proceed by differentiating both sides:
\begin{eqnarray*}
\vx - \frac{1}{2} \nabla\left( \vx^T f(\vx) \right) & = & \nabla \rho(\vx) \\
\vx - \frac{1}{2} \left( f(\vx) + \nabla f(\vx)\:\vx \right) & = & \nabla \rho(\vx) \\
\vx - \frac{1}{2} \left( f(\vx) + f(\vx) \right) & = & \nabla \rho(\vx) \\
\vx - f(\vx) & = & \nabla \rho(\vx)
\end{eqnarray*}
where in the last step we have invoked the directional derivative expression developed in (\ref{eq:Euler3}). The solution, $f(\vx)$, is a denoiser explicitly defined in terms of the prior,
\begin{equation} \label{filterprior}
f(\vx) = \vx - \nabla \rho(\vx).
\end{equation}
This is quite a natural result in retrospect -- it resembles a steepest descent step\footnote{It is interesting to note the resemblance of this expression to a similar expression that holds for proximal operators (see \cite{Proximal}). In this context, the class of denoisers we have described using conditions 1 and 2 is a more general form of proximal mappings.}. To see this, consider the denoising problem regularized by $\rho(\vx)$:
\begin{equation}
\argmin{\vx} \frac{1}{2} \|\vx-\vy\|^2+ \rho(\vx).
\end{equation}
One step of steepest descent with step-size of $1$ and initial condition $\vx_{0}=\vy$, would read
\begin{equation}
\vx_{1} = \vx_{0} - ( \vx_{0}-\vy + \nabla\rho(\vx_{0}) ) = \vy - \nabla\rho(\vy)
\end{equation}
just as shown above as a denoising on $\vy$.

%%%%%%%%%%%%%%%%%%%%%%%%%%%%%%%%%%%%%%%%%%%%%%%%%%%%%%%%%%%%%%%%%%%%%%%%%%%%%%%%
%%%%%%%%%%%%%%%%%%%%%%%%%%%%%%%%%%%%%%%%%%%%%%%%%%%%%%%%%%%%%%%%%%%%%%%%%%%%%%%%

\section{Kernelizing Priors}
We showed in the previous section that a prior with the right properties implies a denoiser beneath; and that this denoiser is directly given by the gradient of the prior.  Next, let's address a related, but more specific question: Given a prior $\rho(\vx)$, does it imply a denoising filter of the {\em pseudo-linear form} $f(\vx) = \bW(\vx) \vx$? These types of filters are of course of great interest because they reveal the kind of weighted averaging done by the denoiser, and they are easy to implement. Given $\vx$, we can compute the weights $\bW(\vx)$ in one step, and apply them as $\bW(\vx) \vx$  to the image in another step. Some of the most popular filters to date, such as bilateral, NLM, and K-SVD, are of this convenient form.

Before we go about finding the hidden filter matrix $\bW(\vx)$, let's illustrate a useful property of the pseudo-linear filters. Take $f(\vx) = \bW(\vx) \vx$, and again invoke the expression $f(\vx) = \nabla f(\vx) \vx $ developed in (\ref{eq:Euler3}). Substitution gives
\begin{equation}
\bW(\vx) \vx = \nabla f(\vx) \vx   \;\;\;\;\;\;\;\;\;\;\ \Longrightarrow \;\;\;\;\;\;\;\;\;\;\ (\nabla f(\vx) - \bW(\vx)) \vx = 0
\end{equation}
\noindent for all $\vx$. From this we conclude that the gradient of pseudo-linear filters is in fact the weight matrix
\begin{equation}
\nabla f(\vx) = \bW(\vx).
\end{equation}

Now we can go after the weight matrix by taking the analysis from the previous section one step further and computing the second derivative of (\ref{eq:priorform}). Starting with the expression (\ref{filterprior}) that arose from the first derivative, we differentiate again,
\begin{eqnarray*}
f(\vx) & = & \vx - \nabla \rho(\vx) \\
\nabla f(\vx) & = & \bI - \nabla(\nabla \rho(\vx)) \\
\nabla f(\vx) & = & \bI - \bH (\rho(\vx)).
\end{eqnarray*}
Replacing $\nabla f(\vx) = \bW(\vx)$, we obtain the pleasing result that the weight matrix implied by the prior is the identity matrix minus the Hessian of the prior,
\begin{equation}
\bW(\vx)  =  \bI - \bH (\rho(\vx)),
\end{equation}
or posed another way, $\bL(\vx) = \bI - \bW(\vx) = \bH (\rho(\vx))$ (i.e. the Laplacian filter is directly given by the  Hessian of the prior, which is not surprising, bearing in mind that we seek a relation of the form $\rho(\vx)= \vx^T \bL(\vx) \vx$).  What we have done here is to ``kernelize'' the regularizer and find an explicit expression for the weights of the implied filter. Several observations about this result are in order:
\begin{itemize}
\item \textbf{Convexity:} If $\rho(\vx)$ is convex, then its Hessian is symmetric positive semi-definite (PSD), and therefore $\bL(\vx)$ is PSD. Furthermore, if $\eta\left(\bL(\vx)\right) \le 1$, i.e., we get that (i) $\bW(\vx)$ has spectral radius equal or smaller than $1$, implying that strong passivity condition (Condition 2) is guaranteed; and (ii) $\bW(\vx) = \bI - \bL(\vx) $ is also PSD -- a desirable property \cite{Symmetric}.

\item \textbf{Homogeneity:} Since $\rho(\vx)$ is 2-homogenous, its gradient is $1$-homogeneous. We can show this by differentiating:
\begin{eqnarray*}
\rho(c\vx) & = & c^2 \rho(\vx)  \\
\nabla_{\vx} \rho(c\vx) & = & c^2 \nabla_{\vx} \rho(\vx) \\
\frac{\partial c\vx}{\partial \vx} \:\nabla_{c\vx} \rho(c\vx) & = & c^2 \nabla_{\vx} \rho(\vx) \\
c\nabla_{c\vx} \rho(c\vx) & = & c^2 \nabla_{\vx} \rho(\vx) \\
\nabla_{c\vx} \rho(c\vx) & = & c \nabla_{\vx} \rho(\vx)
\end{eqnarray*}
Similarly the Hessian $\bH(\rho(\vx))$ is {\em invariant}\footnote{Or $0$-homogenous} to scaling of the image $\vx$, and so is the implied filter matrix $\bW(\vx)$. Consequently, the applied filter $\bW(\vx) \vx$ is $1$-homogenous, which again is consistent with our earlier conditions.
\item \textbf{Row Stochasticity:} Let's recall the expression we derived above for the filter in terms of the Hessian of the prior,
\begin{equation}
\bW(\vx)  =  \bI  -\bH(\rho(\vx)).
\end{equation}
If a filter defined this way is to be row-stochastic, we would have (for every $\vx$),  $\bW(\vx) \mathbb{1} = \left( \bI - \bH(\rho(\vx))\right) \mathbb{1} = \mathbb{1}$, or equivalently,
\begin{equation}
\bH(\rho(\vx))\mathbb{1} = \mathbb{0}.
\end{equation}
This relation does not hold in general. However, consider defining the prior in terms of the gradient of the image instead of the image itself. Namely, this involves a change of variables in the prior $\rho(\vx)$ from $\vx$ to $\bD\vx$, where $\bD$ is the gradient (e.g. difference) operator. For instance, instead of  $\rho(\vx) = \|\vx\|_1^2$, consider  $\rho_\bD(\vx) =  \|\bD\vx\|_1^2$. The Hessian of the prior under this linear transformation is given by the chain rule as
\begin{equation}
\bH(\rho_\bD(\vx))  = \bD^T \bH(\rho(\vx))  \bD.
\end{equation}
This Hessian, when applied to the constant vector $\mathbb{1}$ will vanish for all $\bx$ since $\bD \mathbb{1} = \mathbb{0}$:
\begin{equation}
\bD^T \bH(\rho(\vx))  \bD \:\mathbb{1} = \mathbb{0},
\end{equation}
so the filter resulting from this Hessian is row-stochastic. In fact, the same can be said for column-stochasticness, since $\mathbb{1}^T\bD^T = 0$.
\end{itemize}

%%%%%%%%%%%%%%%%%%%%%%%%%%%%%%%%%%%%%%%%%%%%%%%%%%%%%%%%%%%%%%%%%%%%%%%%%%%%%%%%
%%%%%%%%%%%%%%%%%%%%%%%%%%%%%%%%%%%%%%%%%%%%%%%%%%%%%%%%%%%%%%%%%%%%%%%%%%%%%%%%

\section{More on Homogeneity}

The concept of homogeneity of a filter played an important role in the development of the key results of the paper. So it is worth saying a bit more about it and to question whether this condition is satisfied for some popular and familiar filters.

%%%%%%%%%%%%%%%%%%%%%%%%%%%%%%%%%%%%%%%%%%%%%%%%%%%%%%%%%%%%%%%%%%%%%%%%%%%%%%%%

\subsection{Non-Local Means and Bilateral Filter}

A general construction of a denoising filter could be based on a symmetric positive semi-definite kernel $\bK_{i,j}(\vx) = \bK(x_i,x_j) \geq 0$ from which the filter matrix $\bW(\vx)$ is constructed by normalization. More specifically,
\begin{equation}
\bW_{i,j}(\vx) = \frac{\bK_{i,j}(\vx)}{\sum_{i=1}^n \bK_{i,j}(\vx)}.
\end{equation}
Whether such a denoiser is homogenous very much depends on the choice of the kernel $\bK$. For instance, if the kernel is homogeneous of {\em any} degree $p$, then the resulting filter matrix is invariant to scaling through cancellation,
\begin{equation}
\bW_{i,j}(c\vx) = \frac{\bK_{i,j}(c\vx)}{\sum_{i=1}^n \bK_{i,j}(c\vx)} = \frac{c^p\bK_{i,j}(\vx)}{\sum_{i=1}^n c^p \bK_{i,j}(\vx)} = \bW_{i,j}(\vx).
\end{equation}
Examples of homogeneous kernels include (homogeneous) polynomials, and several others \cite{homokernels} which are not in common use in image processing. Most commonly used kernels are the exponentials (Gaussian to be exact), which are used in the bilateral or non-local means (NLM) cases. The Gaussian function is not homogeneous, but as we will show below, the resulting pseudo-linear filter is nearly so. We will show that for $c = 1+\epsilon$ with very small $\epsilon$ we have $1$-homogeneity for the NLM-type filters, namely:
\begin{equation}
f(c\vx) = \bW(c\bx) (c\vx) = c \bW(\vx) \: \vx  = c f(\vx).
\end{equation}
The $i,j$-th element of the filter weight matrix for the NLM filter\footnote{The bilateral filter, which includes a spatial distance weight can be treated similarly, since the spatial weights are invariant to scaling of the values of the image in any case.} is
\begin{equation}
\bW_{i,j}(\bx;\sigma) = \frac{e_{ij}(\sigma)}{d_j(\sigma)}
\end{equation}
\noindent where
\begin{eqnarray*}
e_{ij} (\sigma)& = & \exp(-\|\bR_i \vx - \bR_j \vx \|^2/2\sigma^2), \\
d_j (\sigma) & = &  \sum_{i=1}^{n} \exp(-\|\bR_i \vx - \bR_j \vx \|^2/2\sigma^2).
\end{eqnarray*}
where $\bR_i \vx$ is a patch centred at pixel position $i$, extracted from the image;  the normalization constant $d_j(\sigma)$ is given by summing across the rows of the kernel matrix.

Do these weights change much when the image $\vx$ is replaced by a scaled version $c\vx$? First, note that if $\sigma$ is nearly zero, then all weights are essentially equal to $1/n$ and therefore they are automatically invariant to scaling of the image. Next, let's consider the other extreme where the value of $\sigma$ is away from zero. Now, note that the scaling in $\vx$ can be absorbed in the parameter $\sigma$ as follows:
\begin{equation}
\exp(-\|c \bR_i \vx - c \bR_j \vx  \|^2/2\sigma^2) = \exp(-\|\bR_i \vx - \bR_j \vx \|^2/2(\sigma/c)^2).
\end{equation}
Second, the effect of this (multiplicative) scaling can be approximated by an additive perturbation,
\begin{equation}
\frac{\sigma}{c} = \frac{\sigma}{1+\epsilon} \approx \sigma-\epsilon \sigma = \sigma+\delta.
\end{equation}
We now compute an approximation to $\bW_{i,j}(\vx,\sigma+\delta)$ using a Taylor series:
\begin{eqnarray*}
\bW_{i,j} ((1+\epsilon)\vx;\sigma) & \approx & \bW_{i,j} (\vx;\sigma+ \delta) \\
 &  \approx &   \bW_{i,j}(\vx;\sigma)  +  \delta \; \frac{\partial \bW_{i,j}(\vx;\sigma)}{\partial \sigma}.
\end{eqnarray*}
The derivative of the weight values will be calculated in terms of the functions $e_{ij}(\sigma)$ and $d_j(\sigma)$ as follows:
\begin{eqnarray*}
\frac{\partial \bW_{i,j}(\vx;\sigma)}{\partial \sigma}  & = & \frac{\partial}{\partial \sigma} \left(\frac{e_{ij}(\sigma)}{d_j(\sigma)} \right) \\
 & = &  \frac{e'_{ij}(\sigma) d_j(\sigma) - e_{ij}(\sigma) d'_j(\sigma)}{d_j^2(\sigma)}  \\
 & = & \frac{e'_{ij}(\sigma)}{d_j(\sigma)}  - \frac{e_{ij}(\sigma)}{d_j(\sigma)} \frac{d'_j(\sigma)}{d_j(\sigma)} \\
 & = & \frac{\|\bR_i \vx - \bR_j \vx \|^2}{\sigma^3}\frac{e_{ij}(\sigma)}{d_j(\sigma)} - \frac{e_{ij}(\sigma)}{d_j(\sigma)} \frac{d'_j(\sigma)}{d_j(\sigma)} \\
& = & \frac{\|\bR_i \vx - \bR_j \vx \|^2}{\sigma^3}\bW_{i,j}(\bx;\sigma) -   \frac{d'_j(\sigma)}{d_j(\sigma)}\bW_{i,j}(\bx;\sigma)\\
& = & \left( \frac{\|\bR_i \vx - \bR_j \vx \|^2}{\sigma^3}-   \frac{d'_j(\sigma)}{d_j(\sigma)} \right) \bW_{i,j}(\bx;\sigma).
\end{eqnarray*}
Therefore,
\begin{equation}
\bW_{i,j} (\vx;\sigma+ \delta)  \approx  \left[ 1 +  \delta \; \left( \frac{\|\bR_i \vx - \bR_j \vx \|^2}{\sigma^3}-   \frac{d'_j(\sigma)}{d_j(\sigma)} \right)\right]  \bW_{i,j}(\bx;\sigma).
\end{equation}
Replacing $\delta = -\epsilon \sigma$, we obtain
\begin{eqnarray*}
\bW_{i,j} ((1+\epsilon) \vx;\sigma) & \approx &  \left[ 1 -\epsilon \; \left( \frac{\|\bR_i \vx - \bR_j \vx \|^2}{\sigma^2}-   \sigma \frac{d'_j(\sigma)}{d_j(\sigma)} \right)\right]  \bW_{i,j}(\bx;\sigma)  \\
& = &   \left( 1 -\epsilon \; \phi(\sigma) \right)  \bW_{i,j}(\bx;\sigma).
\end{eqnarray*}
We can simplify further, but this is not necessary since $\phi(\sigma)$ does not depend on $\epsilon$. For its part, $\phi(\sigma)$ behaves like $n/\sigma^2$. To see this note that the first term in the definition of $\phi$ is at worst $n/\sigma^2$ since $\|\bR_i \vx - \bR_j \vx \|^2$  is bounded by $n$, given that the pixel values are in the range $[0,255]$.

Similarly, $d$ is on the order of $n \cdot \exp(\|\bR_i \vx - \bR_j \vx \|^2/2\sigma^2)$, and its derivative $d'$ is on the order $n \cdot \|\bR_i \vx - \bR_j \vx \|^2 \cdot d / \sigma^3$. Consequently, the second term $ \sigma \cdot d'/d$ behaves like $\|\bR_i \vx - \bR_j \vx \|^2 / \sigma^2$ is also on the order $n/\sigma^2$. Therefore, choosing $\epsilon = 1/n$, for sufficiently large $\sigma$, the term $\epsilon \phi(\sigma)$ becomes negligible.

What we have shown is that the filter weights change very little as a result of the scaling $(1+\epsilon)\vx$ as long as $\epsilon$ is very small. Therefore the NLM (and bilateral) filters are (almost exactly) $1$-homogeneous as we had hoped.

%%%%%%%%%%%%%%%%%%%%%%%%%%%%%%%%%%%%%%%%%%%%%%%%%%%%%%%%%%%%%%%%%%%%%%%%%%%%%%%%

\subsection{Tikhonov Regularizer and Wiener Filtering}

We now turn to show that Tikhonov regularization obeys the homogeneity condition. In this case, the denoised image is the solution of
\begin{eqnarray}
{\widehat \vx} = \argmin{\vx}  \frac{1}{2\sigma^2} \|\vx-\vy\|_2^2 + \frac{\lambda}{2}\|\bB\vx\|_2^2~,
\end{eqnarray}
where $ \bB $, for example, can be a discrete approximation of a derivative operator. The closed-from expression of the above minimization is given by
\begin{eqnarray}
\widehat \vx = f(\vy) & = & \left(\bI+\lambda\sigma^2\bB^T\bB\right)^{-1} \vy \\ \nonumber
                      & = & \bW\vy.
\end{eqnarray}
When we feed this linear denoiser with the scaled image $ (1+\epsilon)\vy $, it gives
\begin{eqnarray} \label{fcy_tikhonov}
f((1+\epsilon)\vy) =\left(\bI+\lambda\sigma^2\bB^T\bB\right)^{-1} (1+\epsilon) \vy,
\end{eqnarray}
which is trivially the same as $ (1+\epsilon)f(\vy) $.

A more challenging case is obtained when the denoiser adapts to the input such that when we apply $ f((1+\epsilon)\vy) $, the denoiser modifies $\lambda $ to be $ \lambda(1+\epsilon)^2 $. In what follows we return to Equation \eqref{fcy_tikhonov}, but this time with the modified $\lambda$, and study the behavior of the filter for infinitesimal change in $ \epsilon $, i.e.,
\begin{eqnarray}
\lim\limits_{\epsilon\rightarrow 0}\bW(\epsilon)  = \lim\limits_{\epsilon\rightarrow 0} \left(\bI+\lambda(1+\epsilon)^2\sigma^2\bB^T\bB\right)^{-1}.
\end{eqnarray}
By relying on the first-order Taylor expansion of the above expression and taking the limit $ \epsilon \rightarrow 0 $ we get the following
\begin{eqnarray}
\bW(\epsilon) & \approx & \bW(0) - \epsilon \ \bW(\epsilon) \left(\lambda(1+\epsilon)^2\sigma^2\bB^T\bB\right)^T \bW(\epsilon) |_{\epsilon = 0} \\ \nonumber
              & = & \bW - \epsilon \ 2 \lambda \sigma^2 \bW \bB^T\bB \bW \\ \nonumber
              & = & (\bI-\epsilon\  2 \lambda \sigma^2 \bW \bB^T\bB)\bW,
\end{eqnarray}
where $ \bW = \bW(0) $. Therefore, we obtain that the adaptive filter $ \bW(\epsilon) $ changes linearly with $ \epsilon $. Moreover, $\| \sigma^2 \bW \bB^T\bB \|<1$, and thus, if $\lambda \ll 1$  the second term becomes negligible (similar to what we have shown for the NLM filter). Therefore, we conclude that the image-adaptive Wiener filtering satisfies the homogeneity condition under mild conditions.

\subsection{Patch-Based Denoising}

We now turn to discuss state-of-the-art patch-based denoising algorithms. These methods clean the input noisy image by (i) breaking it into small overlapping patches, (ii) applying a local denoising step, and (iii) reconstructing the output by averaging the denoised overlapping patches to form the final image. The wisdom of these algorithms relies on the choice of the local non-linear prior. Moreover, in most cases, the denoising process can be divided into two parts; the first contains all the non-linear decisions made by the prior, while the second is nothing but a linear filter that cleans the noisy patches followed by a patch-averaging step. Clearly, the latter fulfills the homogeneity condition due to its linearity. In what follows we argue that the non-linear part of various popular denoisers is stable to an infinitesimal change in scale, thus leading to an overall satisfaction of the homogeneity property.

\subsubsection*{Gaussian Mixture Model}
We start by the EPLL which can be considered as an iterative GMM denoising algorithm. The non-linear part of the GMM prior is the choice of a (pre-trained) Gaussian model that best fits the input noisy patch, and its linear part is the subsequent Wiener filtering step. {Consider a Gaussian model for an $ n $-dimensional patch $ \vx_i \sim N(\textbf{0},\bSigma_k)$, where $ \bSigma_k \in \RR^{n\times n} $ is the $ k $-th Gaussian, taken from the mixture. Following the derivations in \cite{GMM2}, the MAP estimate is formulated by
\begin{eqnarray} \label{wiener_gmm}
{\widehat \vx_i^k} & = & \argmin{\vx_i}  \frac{1}{2} \|\vx_i-\vy_i\|_2^2 + \sigma^2\vx_i^T\bSigma_k^{-1}\vx_i \\ \nonumber
& = & \left(\bI + \sigma^2\bSigma_k^{-1}\right)^{-1}\vy_i \\ \nonumber
& = & \bW_i^k\vy_i,
\end{eqnarray}
which is nothing but the Wiener filter that cleans the $i$-th noisy patch $ \vy_i $. The best model for the $ i $-th patch, $ k_i^* $, is the one that maximizing the MAP over all the possible models, given by
\begin{eqnarray}
k_i^*  = \argmin{k}  \|\widehat \vx_i^k - \vy_i\|_2^2 + \sigma^2(\widehat \vx_i^k)^T\bSigma_k^{-1}(\widehat \vx_i^k) + \sigma^2\log|\bSigma_k|.
\end{eqnarray}
By plugging Equation \eqref{wiener_gmm} into the above we get
\begin{eqnarray} \label{gmm_before_scale}
k_i^* & = & \argmin{k}  \|(\bW_i^k - \bI)\vy_i\|_2^2 + \sigma^2(\bW_i^k\vy_i)^T\bSigma_k^{-1}(\bW_i^k\vy_i) + \sigma^2\log|\bSigma_k| \\ \nonumber
& = &  \argmin{k} \Psi(\vy, \sigma, \bSigma_{k}).
\end{eqnarray}
When we feed the denoiser with $ (1+\epsilon)\vy_i $, the above can be written as
\begin{eqnarray}
k_i^{*,\epsilon} & = & \argmin{k}  (1+\epsilon)^2\|(\bW_i^k - \bI)\vy_i\|_2^2 + (1+\epsilon)^2 \sigma^2(\bW_i^k\vy_i)^T\bSigma_k^{-1}(\bW_i^k\vy_i) + \sigma^2\log|\bSigma_k| \\ \nonumber
& = & \argmin{k} \|(\bW_i^k - \bI)\vy_i\|_2^2 + \sigma^2(\bW_i^k\vy_i)^T\bSigma_k^{-1}(\bW_i^k\vy_i) + \frac{\sigma^2}{(1+\epsilon)^2}\log|\bSigma_k|.
\end{eqnarray}
By relying on the relation $ 1/(1+\epsilon)^2 \approx 1 - 2\epsilon $ we further simplify the above and obtain
\begin{eqnarray}
k_i^{*,\epsilon} & = & \argmin{k}  \|(\bW_i^k - \bI)\vy_i\|_2^2 + \sigma^2(\bW_i^k\vy_i)^T\bSigma_k^{-1}(\bW_i^k\vy_i) + \sigma^2\log|\bSigma_k| - 2\epsilon\sigma^2\log|\bSigma_k| \\ \nonumber
& = & \argmin{k}  \Psi(\vy, \sigma, \bSigma_k) - 2\epsilon\sigma^2\log|\bSigma_k|.
\end{eqnarray}
Now we turn to compare Equation \eqref{gmm_before_scale} to the one derived above, and get the following condition on $ \epsilon $ that guarantees that $ k_i^{*,\epsilon} = k_i^{*} $:
\begin{eqnarray}
\Psi(\vy, \sigma, \bSigma_{k}) - 2\epsilon\sigma^2\log|\bSigma_{k}| > \Psi(\vy, \sigma, \bSigma_{k^*}) - 2\epsilon\sigma^2\log|\bSigma_{k^*}| \\ \nonumber
\end{eqnarray}
Since $ \Psi(\vy, \sigma, \bSigma_{k}) > \Psi(\vy, \sigma, \bSigma_{k^*}) $ one can always choose $ \epsilon \rightarrow 0$ which will keep this inequality intact. In an extremely rare case, when $ \Psi(\vy, \sigma, \bSigma_{k}) = \Psi(\vy, \sigma, \bSigma_{k^*}) $, we can modify the GMM denoiser and propose a simple rule for choosing the model that have a smaller $ \log|\bSigma_{k}| $ term, ensuring that $ k_i^{*,\epsilon} = k_i^{*} $. To conclude, the non-linear part of the GMM denoising algorithm is stable to small scaling of the input.}

\subsubsection*{Sparsity-Inspired Denoisers: K-SVD}
Given a dictionary, the non-linear mechanism of the K-SVD is the Orthogonal Matching Pursuit (OMP) \cite{OMP} algorithm, which estimates the sparse representation of an input noisy patch. {This is a greedy method, aiming to approximate the solution of
\begin{eqnarray}
{\hat \alpha_i} = \argmin{\alpha_i} \|\alpha_i\|_0 \quad \text{s.t.} \quad \| \vy_i - \bD\alpha_i \|_2^2 \leq n\sigma^2,
\end{eqnarray}
where $ n $ is the size of the patch, and $ \bD \in \RR^{n\times m}$ is a (possibly redundant, $ m>n $, and non-orthogonal) dictionary. At each step, denoted by $ k $, the OMP picks a new atom (a column from $ \bD $) that minimizes the residual. Formally, the rule for choosing the first atom $ d_{j_1} $ can be written as
\begin{eqnarray} \label{omp_first}
{j_1} = \argmax{j} |d_j^T\vy_i|,
\end{eqnarray}
while in the $ t $-th step of the OMP, it is the one that maximizing
\begin{eqnarray} \label{omp_rule}
{j_t} = \argmax{j} |d_j^T\vr_i^t|,
\end{eqnarray}
where $ \vr_i^t =  \vy_i - \bD_{S_i^t}\alpha_i^t$ is the residual. We denoted by $ S_i^t $ the set of chosen atoms, obtained in the previous steps, and by $ \bD_{S_i^t} \in \RR^{n\times |{S_i^t}|}$ the corresponding dictionary -- a matrix having the chosen atoms as its columns. This expression can be further simplified by relying on the fact that the representation is the outcome of a least-squares solution, given by
\begin{eqnarray}
\alpha_i^t = \left( \bD_{S_i^t}^T \bD_{S_i^t} \right)^{-1}\bD_{S_i^t}^T\vy_i.
\end{eqnarray}
Substituting the above in Equation \eqref{omp_rule} results in
\begin{eqnarray} \label{omp_subsequent}
{j_t} = \argmax{j} |d_j^T\vr_i| = \argmax{j} \left|d_j^T\left(\bI - \left( \bD_{S_i^t}^T \bD_{S_i^t} \right)^{-1}\bD_{S_i^t}^T\right)\vy_i\right|.
\end{eqnarray}
This process is repeated until reaching to the error constraint.

Trivially, scaling the input patch would not modify the result of Equations \eqref{omp_first} and \eqref{omp_subsequent}. However, scaling the input may modify the stopping rule, and thereby the number of atoms that the OMP picks. This will happen only when $ \|\vy_i - \bD\alpha_i \|_2^2 = n\sigma^2$, which is an extremely rare case. Notice that given the final set of chosen atoms, $ S_i^* $, the cleaned patch is given by $ \hat{\vx}_i = \bD_{S_i^*} \left( \bD_{S_i^*}^T \bD_{S_i^*} \right)^{-1}\bD_{S_i^*}^T\vy_i $, which clearly satisfies the homogeneity condition. To conclude, we showed that the non-linear part of the OMP is stable to an infinitesimal change in scale, and since the cleaned patch is simply obtained by a linear projection onto the chosen atoms we have that with high probability the OMP (and thus the K-SVD) fulfills our hope for homogeneity.
}

%%%%%%%%%%%%%%%%%%%%%%%%%%%%%%%%%%%%%%%%%%%%%%%%%%%%%%%%%%%%%%%%%%%%%%%%%%%%%%%%
%%%%%%%%%%%%%%%%%%%%%%%%%%%%%%%%%%%%%%%%%%%%%%%%%%%%%%%%%%%%%%%%%%%%%%%%%%%%%%%%

\section{The Differentiability Requirement}

In the opening of Section 3.1, we required the denoising engine $f(\vx)$ to be differentiable. Why? There are several benefits for having this behavior:
\begin{enumerate}
\item The directional derivative property, $\nabla_{\vx} f(\vx)\vx=f(\vx)$, which emerges from the homogeneity condition, becomes possible.
\item The passivity condition, which refers to the spectral radius of $\nabla_{\vx} f(\vx)$, stands on solid grounds.
\item The convergence of the fixed-point algorithm discussed in Section 4.2 is guaranteed.
\item The convexity of the regularization term $\rho_L(\vx)$, discussed in Section 5.1, and emerging from Condition 2, becomes possible.
\end{enumerate}
\noindent All these are good reasons for demanding differentiability of the denoiser $f(\vx)$, and yet, the question is whether it is too limiting or whether this requirement could be circumvented.

Observe that Point 1 above could rely on the availability of a far weaker requirement of the availability of all directional derivatives of the form $\nabla_{\vx} f(\vx) \vx$. Indeed, the passivity mentioned in Point 2 could also be posed in terms of a directional derivative, replacing Equation (15) by the somewhat weaker requirement
\begin{eqnarray}\nonumber
\forall \vx \ne 0, ~~~ \frac{\vx^T \nabla_{\vx} f(\vx) \vx}{\vx^T\vx} \le 1,
\end{eqnarray}
under the assumption that $\nabla_{\vx} f(\vx)$ is symmetric.

The question we leave as open at this stage is whether Points 3 and 4 above (convergence of the FP  algorithm and the convexity of our regularization) could rely on the existence of all directional derivatives. At worst, if indeed the denoising engine is not differentiable but has all its directional derivatives, Points 3 and 4 are lost, and the behavior of the proposed algorithm is not clear from a theoretical standpoint.

A different, and more practical question is whether differentiability assumption on $f(\vx)$ is feasible in existing algorithms. The NLM and the Bilateral filter are clearly differentiable. TNRD is differentiable as well since its non-linearities are smooth (linear combination of Gaussian RBF's). EPLL, BM3D, are K-SVD more challenging since their non-linear parts include sharp-decisions. Each of these methods could be $\epsilon$-modified to have a fuzzy decision, thus rendering all of them differentiable with hardly any change in their actual behavior.

\bibliographystyle{ieee}

\end{document}